\documentclass[sigconf]{acmart} 
\AtBeginDocument{%
  }

\acmDOI{10.1145/xxxxxxx.xxxxxxx}

\PassOptionsToPackage{round, authoryear}{natbib} 

\usepackage{natbib} 
\usepackage{float}
\usepackage{amsmath}
\usepackage{amssymb}
\usepackage{multirow} 
\usepackage{placeins}
\usepackage{array}
\usepackage{booktabs}
\usepackage{makecell}
\usepackage{tabularx}
\usepackage{caption}
\usepackage{svg}
\usepackage{graphicx}
\usepackage{subcaption}
\usepackage{enumitem}
\usepackage{xcolor}    
\usepackage{booktabs}  

\usepackage{geometry}

\usepackage{booktabs}       
\usepackage{threeparttable} 
\usepackage{tabularx}       
\usepackage{caption}        
\usepackage{float}          
\usepackage{graphicx}       
\usepackage{multirow}       
\usepackage[table]{xcolor}  
\usepackage{hyperref}
\usepackage{balance}

\definecolor{graybg}{gray}{0.92}

\newcommand{\cmark}{\textcolor{green!60!black}{\ding{51}}} 
\newcommand{\xmark}{\textcolor{red}{\ding{55}}}   
\usepackage{textcomp}

\usepackage{amsmath}

\usepackage{threeparttable}

\usepackage{pifont}  

\usepackage{xeCJK}  
\usepackage{csquotes} 
\MakeOuterQuote{"} 
\usepackage{textcomp}

\usepackage[most]{tcolorbox} 

\definecolor{headerblue}{HTML}{2E3B4E} 

\definecolor{boxheader}{RGB}{50, 60, 75}   
\definecolor{boxbg}{RGB}{250, 250, 250}    
\definecolor{accent}{RGB}{0, 90, 150}      
\definecolor{jsonkey}{RGB}{156, 39, 176}   
\definecolor{jsonstring}{RGB}{56, 142, 60} 
\definecolor{jsonnum}{RGB}{21, 101, 192}   
\definecolor{rowblue}{RGB}{236, 244, 252}

\setlist{nosep} 
\setlist[itemize]{itemsep=4pt, topsep=4pt, leftmargin=*}
\setlist[enumerate]{itemsep=4pt, topsep=4pt, leftmargin=*}
\setlist[description]{itemsep=4pt, topsep=6pt, style=nextline, font=\bfseries\color{accent}}

\newcommand{\sectiontitle}[1]{%
  \vspace{0.4cm}%
  \noindent{\Large\bfseries\color{boxheader} #1}%
  \par\vspace{0.2cm}%
}

\newtcolorbox{promptbox}[1]{
  enhanced,
  breakable,
  colback=boxbg,
  colframe=boxheader,
  colbacktitle=boxheader,
  coltitle=white,
  fonttitle=\bfseries\Huge,
  title={#1},
  arc=3mm,           
  boxrule=1.2pt,     
  drop shadow=black!30!white, 
  left=6mm, right=6mm, top=6mm, bottom=6mm, 
  toptitle=3mm, bottomtitle=3mm, 
}

\newtcolorbox{codebox}{
  enhanced,
  breakable,
  colback=white,      
  colframe=gray!20,   
  boxrule=1pt,
  arc=2mm,
  left=2mm, right=2mm, top=2mm, bottom=2mm,
}

\lstdefinestyle{plainjson}{
  basicstyle=\ttfamily\small,
  columns=fullflexible,
  showstringspaces=false,
  commentstyle=\color{gray}\itshape,
  stringstyle=\color{jsonstring},
  morestring=[b]",
  literate=
   *{0}{{{\color{jsonnum}0}}}{1}
    {1}{{{\color{jsonnum}1}}}{1}
    {2}{{{\color{jsonnum}2}}}{1}
    {3}{{{\color{jsonnum}3}}}{1}
    {4}{{{\color{jsonnum}4}}}{1}
    {5}{{{\color{jsonnum}5}}}{1}
    {6}{{{\color{jsonnum}6}}}{1}
    {7}{{{\color{jsonnum}7}}}{1}
    {8}{{{\color{jsonnum}8}}}{1}
    {9}{{{\color{jsonnum}9}}}{1}
    {:}{{{\color{black}:}}}{1}
    {,}{{{\color{black},}}}{1}
    {\{}{{{\color{black}\{}}}{1}
    {\}}{{{\color{black}\}}}}{1}
    {[}{{{\color{black}[}}}{1}
    {]}{{{\color{black}]}}}{1},
}

\begin{document}

\title{TravelEval: A Comprehensive Benchmarking Framework for Evaluating LLM-Powered Travel Planning Agents}


\author{Weiyi Chen}
\orcid{0009-0006-8742-8668}
\affiliation{%
  \institution{Zhejiang University}
  \city{Hangzhou}
  \country{China}
}
\email{only-chen@foxmail.com}

\author{Shuaixiong Wang}
\affiliation{%
  \institution{Hong Kong Polytechnic University}
  \city{Hong Kong}
  \country{China}
}
\email{2207054220@st.nuc.edu.cn}

\author{Ziyun Gao}
\affiliation{%
  \institution{Zhejiang University}
  \city{Hangzhou}
  \country{China}
}
\email{gaoziyun@zju.edu.cn}

\author{Kaichun Hu}
\affiliation{%
  \institution{Zhejiang University}
  \city{Hangzhou}
  \country{China}
}
\email{tauhkc@zju.edu.cn}

\author{Wangze Ni}
\authornote{Corresponding author.}
\affiliation{%
  \institution{Zhejiang University}
  \city{Hangzhou}
  \country{China}
}
\email{niwangze@zju.edu.cn}

\author{Shimin Di}
\affiliation{%
  \institution{Southeast University}
  \city{Nanjing}
  \country{China}
}
\email{shimin.di@seu.edu.cn}

\author{Chen Jason Zhang}
\affiliation{%
 \institution{Hong Kong Polytechnic University}
 \city{Hong Kong}
 \country{China}
}
\email{jason-c.zhang@polyu.edu.hk}

\author{Lei Chen}
\affiliation{%
  \institution{HKUST (GZ) \& HKUST}
  \city{Guangzhou}
  \country{China}
}
\email{leichen@cse.ust.hk}





\renewcommand{\shortauthors}{Weiyi Chen et al.}

\begin{abstract}
The development of Large Language Models (LLMs) has significantly improved travel planning applications, yet evaluating such models is limited by existing benchmarks' limitations: 1) overemphasis on constraint compliance, neglecting multi‑dimensional qualities like spatio‑temporal cost; 2) datasets lacking real‑world authenticity and coverage in key areas (e.g., lodging, transport); and 3) isolated daily plan assessments that miss critical details (e.g., the impact of daily accommodation and visit pacing) needed for entire plan's evaluation. To address this gap, we introduce \texttt{TravelEval}, a realistic and comprehensive benchmark. \texttt{TravelEval} features 1) a novel six-dimensional evaluation framework to holistically assess plans across accuracy, compliance, temporality, spatiality, economy, and utility dimensions; 2) a highly realistic data sandbox with precise accommodation pricing and authentic intercity transportation data; and 3) a simulation-based global evaluation method that emulates complete travel plans with API-integrated geographic information and fine-grained queuing time.
Evaluating 12 mainstream approaches with \texttt{TravelEval} reveals several valuable insights, such that LLMs struggle with globally‑optimized multi‑dimensional planning (especially in spatio‑temporal reasoning and budget compliance), and agentic reasoning strategies offer no consistent improvement. Concisely, \texttt{TravelEval} facilitates travel plan evaluation via grounded spatio-temporal emulation and comprehensive metrics, providing a robust foundation for advancing LLM-powered travel planning research and applications.\footnote{The data and code are available at https://github.com/onlycwy11/TravelEval}

\end{abstract}






\maketitle

\balance

\begin{figure}[H]
\vspace{-2pt}
    \centering
    \includegraphics[width=1\linewidth]{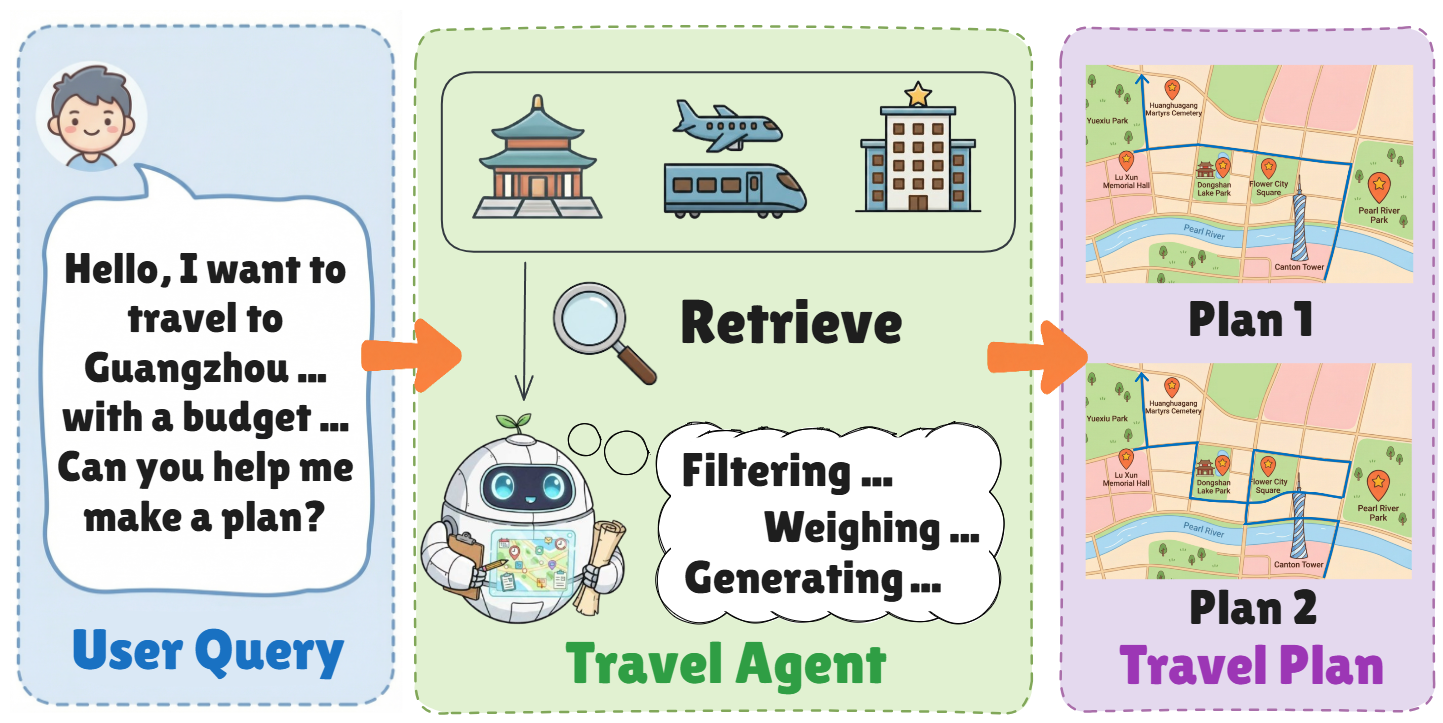}
    \caption{How travel plans generated by an agent}
    \label{fig:travelplanagent}
    \vspace{-4pt}

\end{figure}

\begin{table*}[h]
\centering
\small 
\caption{Comparison with existing travel planning benchmarks across key evaluation dimensions. The dimensions include Accuracy (Acc), Compliance (Com), Temporality (Tem), Spatiality (Spa), Economy (Eco) and Utility (Uti). The sandbox covers Attraction Categories (AC) and Hotel Room Types (HRT). The evaluation methods incorporate Queuing Time (QT), Route Map API (RMA), Attraction Daily Density with Groups (ADD) and Accommodation Spatial Integration (ASI). \cmark~and \xmark~indicate the presence and absence of each capability.}
\label{tab:comparison}

\begin{tabular*}{\textwidth}{@{\extracolsep{\fill}}l cccccc cc cccc@{}}
\toprule
& \multicolumn{6}{c}{\textbf{Dimension}} 
& \multicolumn{2}{c}{\textbf{Sandbox}}  
& \multicolumn{4}{c}{\textbf{Evaluation Method}}\\ 

\cmidrule(lr){2-7} \cmidrule(lr){8-9} \cmidrule(l){10-13}

\textbf{Work} & \textbf{Acc} & \textbf{Com} & \textbf{Tem} & \textbf{Spa} & \textbf{Eco} & \textbf{Uti} & \textbf{AC} & \textbf{HRT} & \textbf{QT} & \textbf{RMA} &\textbf{ADD} &\textbf{ASI}\\ 
\midrule

TravelPlanner~\cite{xie2024travelplanner}       & \cmark & \cmark & \xmark & \xmark & \xmark & \xmark & \cmark & \xmark & \xmark & \cmark & \xmark & \xmark\\
ChinaTravel~\cite{shao2024chinatravel}         & \cmark & \cmark & \xmark & \xmark & \xmark & \xmark & \cmark & \xmark & \xmark & \xmark & \xmark & \xmark\\
Flex-TravelPlanner~\cite{oh2025flextravelplannerbenchmarkflexibleplanning}  & \xmark & \cmark & \xmark & \xmark & \xmark & \xmark & \xmark & \xmark & \xmark & \xmark & \xmark & \xmark\\
TripScore~\cite{qu2025tripscorebenchmarkingrewardingrealworld}           & \cmark & \cmark & \xmark & \xmark & \xmark & \xmark & \cmark & \xmark & \xmark & \xmark & \xmark & \xmark\\
TP-RAG~\cite{Ni2025TPRAG}              & \cmark & \cmark & \cmark & \cmark & \xmark & \cmark & \xmark & \xmark & \xmark & \xmark & \xmark & \xmark\\
TravelAgent~\cite{chen2024travelagent}         & \cmark & \cmark & \xmark & \xmark & \xmark & \xmark & \xmark & \xmark & \xmark & \cmark & \xmark & \xmark\\
TripTailor~\cite{wang2025triptailor}        & \cmark & \cmark & \xmark & \cmark & \xmark & \cmark & \cmark & \xmark & \xmark & \xmark & \xmark & \cmark\\
TripCraft~\cite{chaudhuri2025tripcraftbenchmarkspatiotemporallyfine}           & \cmark & \cmark & \cmark & \cmark & \xmark & \cmark & \cmark & \textbf{Non-Public} & \xmark & \cmark & \cmark & \cmark\\
\midrule
\textbf{TravelEval (Ours)} & \cmark & \cmark & \cmark & \cmark & \cmark & \cmark & \cmark & \cmark & \cmark & \cmark & \cmark & \cmark\\ 
\bottomrule
\end{tabular*}
\end{table*}

\section{Introduction}
\label{sec:introduction}




According to the World Travel \& Tourism Council (WTTC), travel and tourism contributed US\$10.9 trillion to the global GDP in 2024, accounting for 10\% of the global economy and supporting 357 million jobs worldwide~\cite{wttc_economic_impact}. With the development of LLM-powered agents, their application in travel planning has attracted widespread attention from academia (e.g., the KDD community ~\cite{quert2023}) and industry (e.g., ItiNera ~\cite{tang-etal-2024-itinera}). However, as a new paradigm, the quality of the generated travel plans remains unknown and requires a comprehensive evaluation.

As illustrated in Figure~\ref{fig:travelplanagent}, in the emerging paradigm of LLM-powered travel agents, users only need to specify travel needs in natural language, and then the agent will autonomously analyze and call tools to generate an executable travel plan~\cite{chen2024travelagent}. The agent first parses the query to extract key constraints and infers implicit requirements. It then invokes relevant APIs to retrieve feasible options, filters them by availability and cost, and weighs their compatibility with the user's preferences and constraints. Finally, the agent generates a coherent plan that is consistent with the user's constraints and preferences.    


While travel planning agents are evolving rapidly, research on comprehensively evaluating travel planning agents remains nascent. Existing evaluation frameworks like TravelPlanner~\cite{xie2024travelplanner} and ChinaTravel~\cite{shao2024chinatravel} focus their evaluations on whether the model can satisfy all preset hard and soft constraints (e.g., budget, time, transportation methods), or seeking optimal values of single metrics (e.g., minimizing travel time or maximizing the number of attractions). 
However, constraint satisfaction is not equivalent to itinerary quality. Even when all constraints are met, the resulting plan can still be poorly organized, with terrible pacing and unnecessary spatiotemporal overhead. As shown in Figure~\ref{fig:travelplanagent}, both Plan~1 and Plan~2 comply with the strict constraint, yet Plan~2 incurs a substantially higher spatiotemporal cost due to inferior route organization. In other words, a genuinely capable travel planning agent should generate itineraries that are feasible and close-to-optimal across multiple quality dimensions (e.g., arranging attractions in a logical route rather than zigzagging back and forth across the city), rather than merely passing constraint checks~\cite{wang2025triptailor}.

Thus, we need to solve a technical question: how to design a benchmark for comprehensively evaluating LLM-powered agents' travel plan generation ability? However, the question is difficult, and three challenges need to be solved:


\begin{itemize}
    \item \textbf{Challenge 1. Multifarious and intertwined objectives.} Travel planning is a multi-objective optimization problem with intertwined objectives, making it hard to identify and distinguish the evaluation dimensions and design metrics.
    \item \textbf{Challenge 2. Multi-dimensional fine-grained data acquisition.} 
    When evaluation extends from the single dimension of constraint compliance to multiple quality dimensions of travel planning, more accurate multi-dimensional data (e.g., precise transport schedules, hotel room numbers and prices) is required, increasing the difficulty of evaluation data preparation.
    \item \textbf{Challenge 3. Interdependent Events.} Travel planning involves interdependent events. For example, the location of the hotel checked into the previous night impacts the travel distance to the first attraction the next day. Thus, the execution performance of a travel plan cannot be simply measured by the sum of individual event performances, but must account for inter-event mutual effects, which complicates the evaluation.
\end{itemize}

In this paper, we propose \texttt{TravelEval} to address the aforementioned challenges and make the contributions:

\begin{itemize}
    \item  \underline{To address Challenge 1}, we conduct a comprehensive review of relevant papers and extensive practical cases, followed by in-depth analysis. Considering the characteristics of LLMs (e.g., hallucinations) and user needs in travel scenarios (e.g., attraction preferences), we identify six orthogonal evaluation dimensions, including Accuracy, Compliance, Temporality, Spatiality, Economy, and Utility, with 22 novel metrics designed. (\S~\ref{sec:metrics})
    \item  \underline{To overcome Challenge 2}, based on the proposed six evaluation dimensions, we analyzed data requirements and acquired various precise data via multiple APIs to augment the existing dataset. This effort resulted in a dataset with 1,150 user queries and a data sandbox integrating 10 cities, 19,584 POIs, 4,976 accommodations, and 4,655 restaurants. (\S~\ref{sec:data_collection})
    \item  \underline{To solve Challenge 3}, we propose a novel simulation-based evaluation method that simulates the execution of complete travel plans in a sandbox environment, considering the impacts of queuing time, transportation overhead, and attraction density on their execution (e.g., traffic issues may prevent subsequent plans from being executed as scheduled). (\S~\ref{sec:evaluation_methods})
    \item  \underline{For practical evaluation}, we benchmark agent based on 7 leading LLMs with 4 mainstream reasoning strategies and reveal critical insights in terms of feasibility, reasoning strategy efficiency, and task scalability. Notably, regarding task scalability under high complexity, we observe a non-linear performance collapse as user preferences accumulate, where agents tend to prioritize rigid numerical constraints (e.g., budget) at the steep cost of user preference alignment. (\S~\ref{sec:ExperimentalEvaluation})
\end{itemize}

Further, \S~\ref{sec:relatedwork} reviews related work and \S~\ref{sec:conclusion} concludes the paper.
\section{Related Work}
\label{sec:relatedwork}


Recent advances in Large Language Models (LLMs) have extended their evaluation and application across diverse professional domains, including AI-assisted scientific peer review ~\cite{Wang2026WhenAR}, automated code service transformation ~\cite{Ouyang2025Code2MCPTC}, unified tool interfaces for open-source agent collaboration ~\cite{DiToolRosettaBO}, multi-agent systems for engineering simulation automation ~\cite{pan2026automating}, and intelligent material discovery platforms ~\cite{wang2026peromas}. These efforts collectively demonstrate the versatility of LLMs in tackling complex, domain-specific tasks while highlighting the growing need for systematic assessment of their real-world reliability and performance.

\textbf{Methods for Travel Planning.}
Researchers have developed diverse approaches to address travel planning challenges. Specifically, in the realm of automated travel planning, current research explores LLMs as the backbone for autonomous agents~\cite{wang2024survey}, leading to a paradigm shift from combinatorial optimization methods~\cite{gavalas2014survey} and collaborative filtering-based recommendation systems~\cite{lim2015personalized} to language-driven autonomous reasoning. Within this paradigm, existing works integrate critical capabilities, such as multi-step reasoning~\cite{Kojima2022LargeLM} and advanced strategic prompting~\cite{yao2023reactsynergizingreasoningacting, shinn2023reflexionlanguageagentsverbal}, leveraging powerful language understanding to guide LLMs in generating travel itineraries. Subsequent studies further incorporate the utilization of external tools~\cite{schick2023toolformer, qin2023toolllm}, enabling agents to address complex execution requirements beyond pure text generation.


\textbf{Benchmarks for Travel Planning.} Recent research on travel planning has seen significant diversification in benchmark development. One major strand has evolved from assessing itinerary feasibility under fixed constraints (e.g., TravelPlanner~\cite{xie2024travelplanner}) to evaluating multi-turn, flexible reasoning under changing and prioritized constraints (e.g., Flex-TravelPlanner~\cite{oh2025flextravelplannerbenchmarkflexibleplanning} and TripScore~\cite{qu2025tripscorebenchmarkingrewardingrealworld}). 
Another strand focuses on comprehensive and detailed assessment frameworks. TripCraft~\cite{chaudhuri2025tripcraftbenchmarkspatiotemporallyfine} advances evaluation granularity by introducing continuous metrics for LLM-generated travel plans, while TravelAgent~\cite{chen2024travelagent} extends rationality-based criteria with comprehensiveness and personalization. Furthermore, to enhance real-world relevance, recent benchmarks have utilized datasets featuring large-scale POIs and user queries close to real-world travel scenarios~\cite{shao2024chinatravel, wang2025triptailor}. However, as shown in Table~\ref{tab:comparison}, most travel planning benchmarks provide limited coverage of critical evaluation dimensions, mainly focusing on accuracy and compliance while overlooking temporality, spatiality, economy, and utility, as well as lacking authentic real-world coverage in key aspects and relying on per-day independent evaluation of travel plans. To bridge this gap, we propose a framework that integrates multi-dimensional metrics with a realistic sandbox and a simulation-based evaluation method. 

\section{The \texttt{TravelEval} Benchmark}
\label{sec:travelevalBenchmark}



\begin{figure*}
    \centering
    \includegraphics[width=\textwidth]{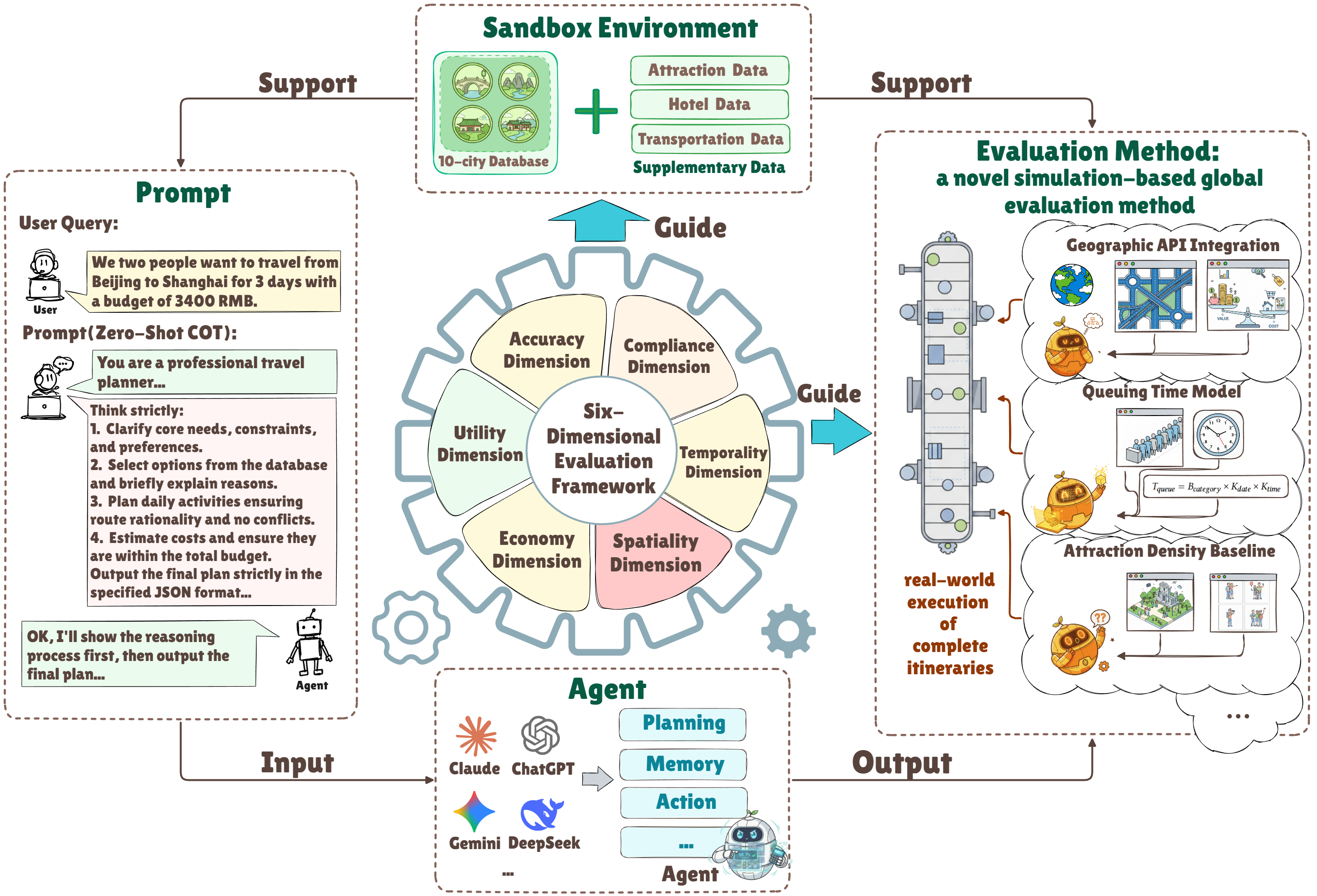}
    \caption{Overview of the \texttt{TravelEval} framework.}
    \label{fig:overview}
\end{figure*}

We propose \texttt{TravelEval}, a benchmark for evaluating LLMs' travel planning capabilities under multi-dimensional and complex constraints (Figure~\ref{fig:overview}). To achieve fine-grained evaluation, \texttt{TravelEval} is comprised of three core components designed to work in concert: (1) a novel \textbf{Six-dimensional Evaluation Framework} that enables comprehensive assessment of travel planning; (2) a highly realistic \textbf{Data Sandbox} characterized by refined and accurate data; and (3) a simulation-based global \textbf{Evaluation Method} leveraging spatial-temporal models. 

\subsection{Six-Dimensional Evaluation Framework}
\label{sec:metrics}


To address Challenge 1 and move beyond the one-sided evaluation of travel planning agents, we conduct exhaustive literature reviews~\cite{Chang2024survey10.1145/3641289, DBLP:journals/electronicmarkets/HleeYL25, li2025exposingnumeracygaps, routeplanningmodel} and in-depth analysis of real-world travel planning cases. As discussed in \S~\ref{sec:introduction}, travel planning agent capability does not merely equate to satisfying explicit constraints; a high-quality itinerary must also exhibit spatiotemporal rationality, balanced economic costs, and strong user experience utility. Building on this insight, and informed by the inherent characteristics of LLMs (e.g., hallucinations and stochasticity) and the multi-faceted nature of travel planning (e.g., spatiotemporal rationality and user experience demands), we induce a six-dimensional evaluation framework and design normalized, quantifiable metrics for each dimension, covering not only the factual accuracy and constraint satisfaction of generated plans but also their spatiotemporal rationality, economic balance, and user experience utility.


\textbf{Accuracy}. 
Given that studies and practices have shown LLMs are insensitive to numbers, prone to miscalculations~\cite{li2025exposingnumeracygaps}, and may fabricate attractions or misestimate intercity travel durations, this dimension evaluates the authenticity and computational reliability of the system output information, which is the basis of planning feasibility, including: \textit{\textbf{(1) Cost Calculation Deviation (CCD)}:} The arithmetic deviation in calculating various costs to measure the LLM's computational capability. 
 \textit{\textbf{(2) Fictitious Attraction Rate (FAR)}:} The proportion of unreal or unverifiable attractions in sandbox recommended by the LLM, measuring its anti-hallucination capability. \textit{\textbf{(3) Violation Rate of Opening Hours (VROH)}:} The proportion of attraction visits that occur outside the official opening hours, measuring the model's temporal perception capability. \textit{\textbf{(4) Failure Rate of Traffic Connection (FRTC)}:} The percentage of traffic transfer failures between consecutive traffic segments due to time/shift conflicts, which measures the model's traffic planning capability. \textit{\textbf{(5) Intercity Traffic Deviation (ITD)}:} The absolute relative deviation between the planned and actual intercity transportation duration to measure the LLM's time estimation capability. \textit{\textbf{(6) Personnel Deviation (PD)}:} The relative deviation between the allocated headcount in various services (e.g., accommodation, transport) and the user-specified number, measuring the model's ability to perceive and apply travel party size consistently.

\textbf{Compliance}. 
To align with mainstream travel planning benchmarks that all incorporate compliance evaluation, we inherit core compliance metrics and propose novel ones (marked with *) to refine fine-grained constraint compliance assessment. This dimension evaluates whether a plan meets the hard constraints specified by users, including: \textit{\textbf{(1) Budget Compliance Sign (BCS)}:} Indicate whether total travel cost is strictly within the user-set budget, measuring the model's budget adherence ability. \textit{\textbf{(2) Time Compliance Sign (TCS)*}:} Verify whether travel days and start/end time fully comply with the user-specified information. \textit{\textbf{(3) Headcount Compliance Sign (HCS)*}:} Check whether planned capacity of activities, transportation, accommodation, and other resources meets the number of users, measuring the model's headcount control precision. \textit{\textbf{(4) Preference Achievement Sign (PAS)}:} Indicate whether accommodation type (A), transportation mode (T) and cuisine preference (C) are configured according to user requirements, measuring the model's preference recognition capability. \textit{\textbf{(5) Tour-Time Compliance Sign (TTCS)*}:} Determine whether the attraction tour duration strictly falls within sandbox-recommended range, measuring the model's duration optimization capability.

\textbf{Temporality}. 
Given that time utilization is a critical factor influencing the quality of travel planning~\cite{DBLP:journals/electronicmarkets/HleeYL25, routeplanningmodel}, this dimension evaluates the utilization efficiency of time resources in travel planning, including: \textit{\textbf{(1) Scene-Time Ratio (STR)}:} The ratio of effective tour time to total allocated time for a single attraction (including queuing), measuring the model's single-attraction time allocation capability. \textit{\textbf{(2) Daily Time Utilization (DTU)}:} The proportion of effective attraction visit duration throughout the day from hotel departure, measuring the model's daily schedule optimization capability. \textit{\textbf{(3) Overall Time Utilization (OTU)}:} The proportion of effective visit duration of all attractions in the total travel duration, measuring the model's global time utilization capability. 

\textbf{Spatiality}.
As spatial rationality is a core determinant of travel experience and itinerary feasibility~\cite{DBLP:journals/electronicmarkets/HleeYL25, ZHONG2024103852SCFL}, this dimension evaluates the logical rationality of the geographical distribution of attractions and access routes, including: \textit{\textbf{(1) Scene Sequence Rationality (SSR)}:} The relative deviation between the actual and theoretical optimal route distances (based on the real road network) of consecutive POIs, measuring the model's route optimization capability. \textit{\textbf{(2) Cross-Day Spatial Mismatch (CSM)}:} The overall spatial mismatches of a multi-day trip characterized by the 90th and 95th percentiles (P90, P95) of CSM values across all attractions, measuring the model's spatial clustering capability. 


\textbf{Economy}. 
As empirical analysis shows that efficient budget-to-experience conversion improves travel planning quality~\cite{routeplanningmodel}, this dimension evaluates the rationality of budget allocation, including: \textit{\textbf{(1) Budget Efficiency (BE)}:} The ratio of travel experience value to actual total expenditure, measuring the model's budget-value conversion capability. \textit{\textbf{(2) Cost Composition}:} The proportion of the total cost for various categories (such as attractions, accommodations, intercity and intracity transportation and meals), measuring the model's budget structure optimization capability.

\textbf{Utility}. 
As recommendation algorithms tailor personalized plans to the popularity of POIs and user preferences/constraints~\cite{ESPER2025+tour}, this dimension is the top level of the evaluation framework quantifying tourists' comprehensive travel planning experience, including: \textit{\textbf{(1) Experience Diversity Index (EDI)}:} Adapted from the Shannon-Wiener Index~\cite{Shannon-Wiener2013} in ecology, measuring the richness and distribution balance of the experience types covered during the trip (e.g., Instagram-Worthy Locations), evaluating the model's experience diversification capability. \textit{\textbf{(2) Attractions Density Score (ADS)}:} Evaluate the alignment between daily attraction quantity and user expectations, measuring the model's demand-aligned density adjustment capability. \textit{\textbf{(3) Attractions Quality Efficiency (AQE)}:} Evaluate whether the LLM allocates more time to high-quality (i.e., high-score) attractions, measuring its quality-prioritized time allocation capability. \textit{\textbf{(4) Profit}:} Quantify the preference-weighted average experience value per attraction, measuring the model's preference-aligned experience optimization capability. 

In Appendix \ref{sec:Appendix_A}, we provide the formal mathematical definitions and detailed logic for the six-dimensional evaluation metrics.

\subsection{Benchmark Construction}
\label{sec:data_collection}
The development of our multi-dimensional evaluation framework imposes specific requirements for comprehensive, targeted data to support its layered assessment logic. In this section, we detail the collection of sandbox data and the generation of user queries, both of which establish a realistic, reliable, and reproducible testing environment for evaluating LLMs' travel planning capabilities.

\textbf{Sandbox Environment Establishment.}
Our sandbox is built upon the ChinaTravel dataset, the official benchmark of~\href{https://chinatravel-competition.github.io/IJCAI2025/}{the IJCAI 2025 Travel Planning Challenge}, which provides a robust, community-validated foundation for real-world travel planning evaluation. However, it exhibits inherent limitations: overly basic and undergranular attribute schemas, ambiguous field definitions, incomplete geographic coverage (notably for granular categories like city-wide hotel inventories), and non-official sourcing for some data streams. To enhance its realism and evaluative capability, we have made vital refinements and additions:

\begin{itemize}[leftmargin=*]
    \item \textbf{Attraction Data.} To capture the nuanced preferences of modern travelers, we first curated a high-quality foundation by scraping attraction ratings from mainstream platforms including Dianping, and then introduced a novel dual-layer classification system (14 foundational and 10 trending types) to systematically characterize attraction attributes. Leveraging prompt engineering, we employed LLMs to automate the initial classification of attractions, with the resulting annotations validated via a double-annotator cross-validation strategy that demonstrated high inter-rater consistency and robust annotation quality.

    \item \textbf{Hotel Data.} We expanded the dataset and refined its schema by replacing ambiguous attributes (e.g.,~\textquotedbl price\textquotedbl, \textquotedbl numbed\textquotedbl) with concrete room-type details (e.g., single, twin, family). To ground pricing in reality, we developed a city-level price coefficient matrix via systematic surveys across 10 cities, and implemented inventory allocation rules to simulate constrained resources.
    
    \item \textbf{Transportation Data.} To emulate realistic travel scenarios, we upgraded the transportation data of 10 cities with real high-speed rail and flight details, and further integrated the Amap API to supplement geographic and traffic information.
\end{itemize}

\textbf{User Query}. 
We constructed a high-quality dataset of 1,150 user queries for rigorous evaluation of travel planning agents, drawing on difficulty stratification methodologies from existing travel planning benchmarks (e.g., TravelPlanner) for empirical alignment and comparability. It is partitioned into two categories: \textbf{Group A: Independent Queries}. This core test set is stratified by difficulty: \textit{easy} (essential constraints only), \textit{medium} (1--2 preferences), and \textit{hard} (3--4 complex preferences including exclusions). To simulate budget constraints, 30\% of Hard queries are assigned a tight~\textquotedbl economy\textquotedbl~budget.
\textbf{Group B: Progressive Groups}. This subset is designed to isolate the impact of evolving user demands on agent performance: all queries in a group share identical core parameters, with only user preferences and constraints varying across three difficulty levels (G1, G2, G3 denote the \textit{easy}, \textit{medium} and \textit{hard} tiers defined in Group A), allowing for controlled evaluation of an agent's sensitivity to evolving demands. All queries were generated by LLMs under strict template rules and manually reviewed by human editors, resulting in a high-quality, consistent query set that closely mirrors real-world travel planning requests.

\subsection{Enhanced Evaluation Methods}
\label{sec:evaluation_methods}
Driven by the demands of our six-dimensional evaluation framework, we propose a novel simulation-based global evaluation method. This method holistically captures the real-world execution of complete itineraries by accounting for critical travel factors that shape plan feasibility and quality, including transportation overhead, queuing time, and attraction density.

\textbf{Geographic API Integration}.
For the \textit{Spatiality} dimension in \S~\ref{sec:metrics}, we integrate the Amap API to retrieve real road network distances between POIs, replacing geodesic distance calculations to ensure spatial metrics reflect authentic travel routing costs.
    
\textbf{Queuing Time Model}. Queuing time is a critical factor for realistic travel planning yet is absent from public evaluation methods. To address this, we introduce a data-driven queuing time model. Its foundation is a multi-variable formula: $T_{queue} = B_{category} \times K_{date} \times K_{time}$,
where $B_{category}$ denotes the category-specific base queuing time, while $K_{date}$ and $K_{time}$ are the date-dependent and time-period adjustment factors. This model provides a realistic time-cost estimate for attraction visits, enabling accurate quantitative evaluation of the \textit{Temporality} dimension within our framework.

\textbf{Attraction Density Baseline.} To ground the \textit{ADS} (in \textit{Utility}) in real-world travel behaviors, we collected and analyzed travelogues across four prevalent travel types (slow travel, family tour, regular tour, and special forces travel) from platforms like Mafengwo and Xiaohongshu, and established reference values by calculating the median daily attraction count for each type, which serves as a realistic reference for evaluating travel pace. 



\section{Experimental Evaluation}
\label{sec:ExperimentalEvaluation}
In this section, we seek to 1) systematically evaluate LLMs' travel planning capabilities against the six-dimensional evaluation metrics in \texttt{TravelEval}, and further disentangle the respective impacts of model architectures and prompting strategies on plan quality; 2) explore performance variations of LLMs in response to changes in core parameters and user preferences; 3) independently examine the discriminative power of progressive difficulty settings driven by the escalation of such preferences for different model-strategy combinations; and 4) investigate the evolution of travel planning performance across successive LLM generations from the same vendor. Accordingly, we conduct extensive experiments on real-world datasets to answer the following research questions.

\begin{itemize}
    \item \textbf{Q1.} How do representative LLMs perform on the six evaluation dimensions defined in \texttt{TravelEval}, and to what extent do these performances vary with (a) different LLMs (fixed prompt) or (b) different prompts (fixed LLM)?
    
    
    
    \item \textbf{Q2.} How does the six-dimensional performance of LLMs change as travel planning tasks increase from easy to hard?

    \item \textbf{Q3.} How does the progressive difficulty setting differentiate the behavior of different model-strategy combinations?

    \item \textbf{Q4.} How do successive generations of LLMs impact travel planning performance across the six evaluation dimensions?
     
    \item \textbf{Q5.} Can external tool calling be effectively integrated into travel planning?
\end{itemize}

\subsection{Experimental Setup}

\textbf{\underline{Models and Prompting Strategies}}. While API access for various commercial agent platforms is currently restricted, our benchmark supports both LLM and agent evaluations. We primarily select representative LLMs for experiments, alongside a locally deployed agent for validation. To evaluate the differences in travel planning performance across model types and mainstream prompting strategies, we conduct benchmark evaluations on \texttt{TravelEval} for 7 representative LLMs. These include both closed-source and open-source models, namely \href{https://api-docs.deepseek.com/news/news250922}{DeepSeek-V3.1}, \href{https://developers.googleblog.com/en/gemini-2-family-expands/}{Gemini-2.0-Flash}, \href{https://openai.com/api/}{OpenAI GPT-4o}, \href{https://openai.com/api/}{OpenAI GPT-4o mini}, \href{https://openai.com/api/}{OpenAI GPT-5-chat}, \href{https://www.anthropic.com/news/claude-sonnet-4-5}{Claude-Sonnet-4.5}, and \href{https://qwen.ai/blog?id=qwen3}{Qwen3-8B}. To further validate \texttt{TravelEval}'s applicability to agentic systems, we additionally incorporate a locally deployed tool-augmented agent. The tests are carried out using four mainstream Prompting Strategies, including Direct (D) reasoning, Zero-Shot Chain-of-Thought (Zero-shot CoT, ZS)~\cite{wei2023chainofthoughtpromptingelicitsreasoning}, ReAct (RA)~\cite{yao2023reactsynergizingreasoningacting}, and Reflexion (Re)~\cite{shinn2023reflexionlanguageagentsverbal}. 


\textbf{\underline{Baseline}}. To establish a reliable reference for assessing LLM-generated travel planning performance, we propose a travel planning generation method based on code logic as the evaluation baseline. This method is designed with two different approaches for different user scenarios: \textbf{Approach A} allows travel plans to moderately exceed the budget, focusing on maximizing users' attraction experience and meeting user preferences as much as possible; \textbf{Approach B} strictly follows budget constraints and ensures that all attractions are accessed in compliance during opening hours. 

Full implementation details are provided in Appendix~\ref{sec:Appendix_A}.

\subsection{Main Results and Analysis}
\label{sec:MainResults}

\begin{figure}[H]
    \centering
    \includegraphics[width= 0.48\textwidth]{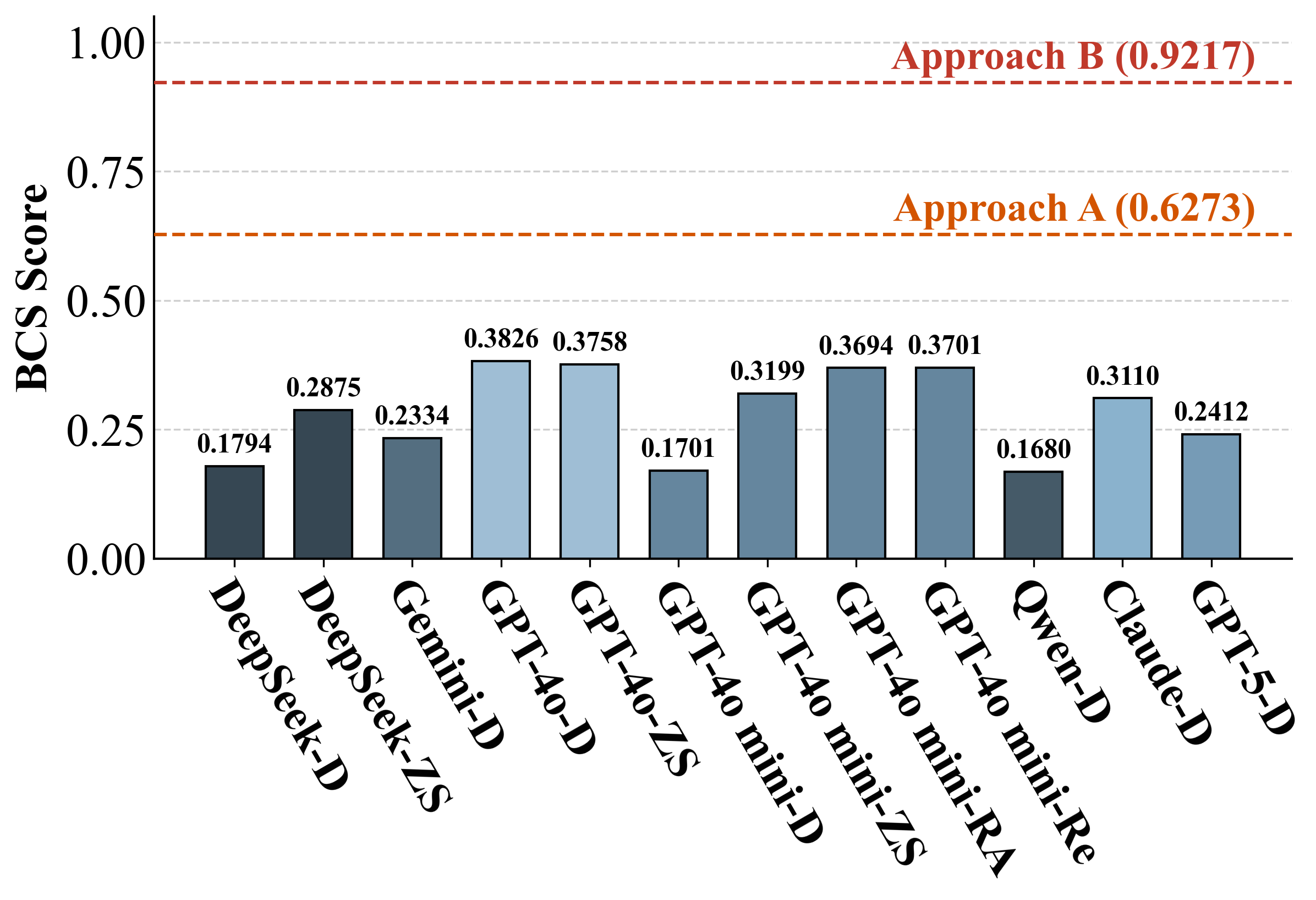}
    \caption{All performances in BCS.}
    \label{fig:BCS_performances}
\end{figure}
\begin{figure*}[t]
    \centering
    \begin{subfigure}[b]{0.48\textwidth}
        \centering
        \includegraphics[width=\textwidth]{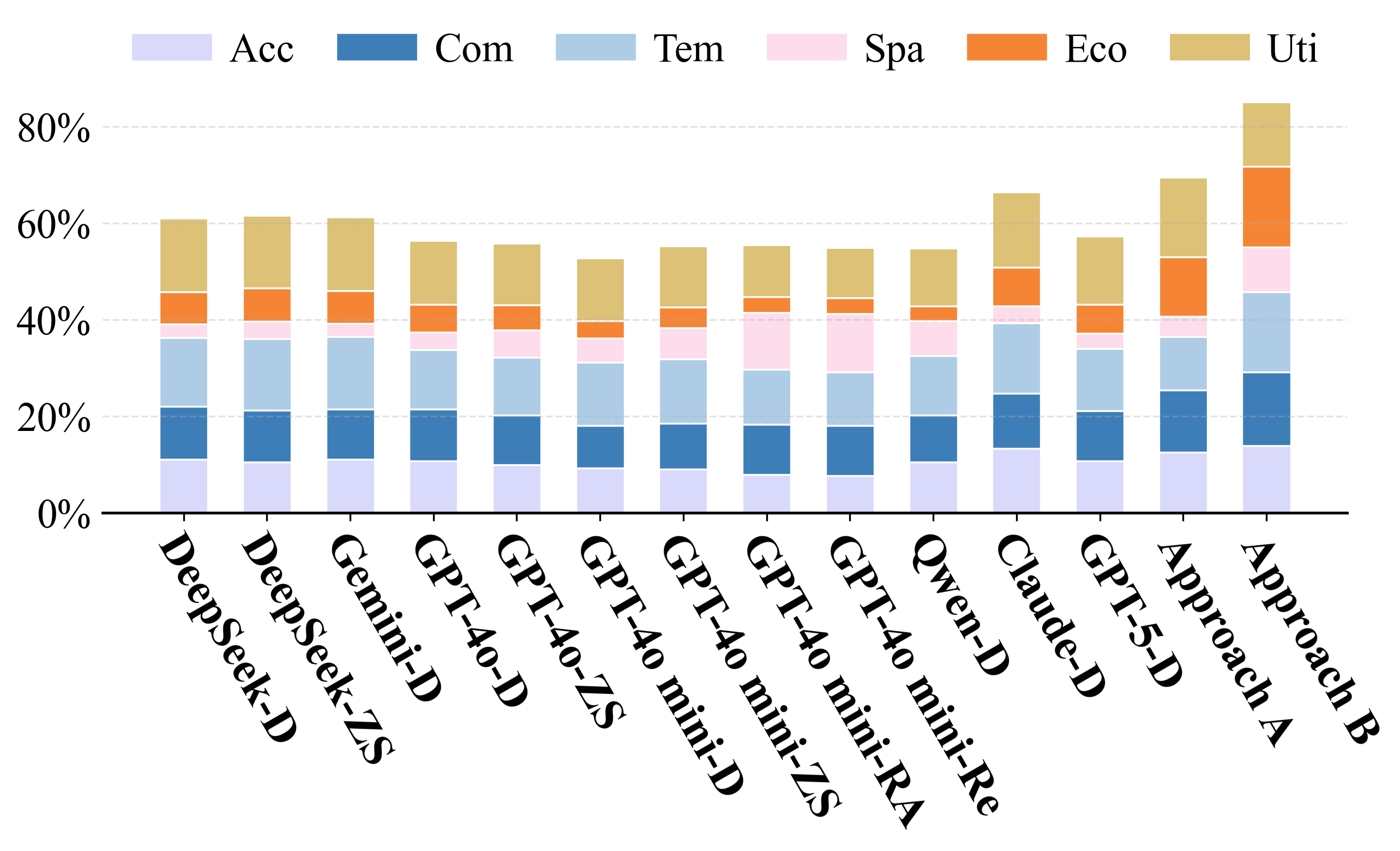}
        \caption{Overall Dataset} 
        \label{fig:overall}
    \end{subfigure}
    \hfill 
    \begin{subfigure}[b]{0.48\textwidth}
        \centering
        \includegraphics[width=\textwidth]{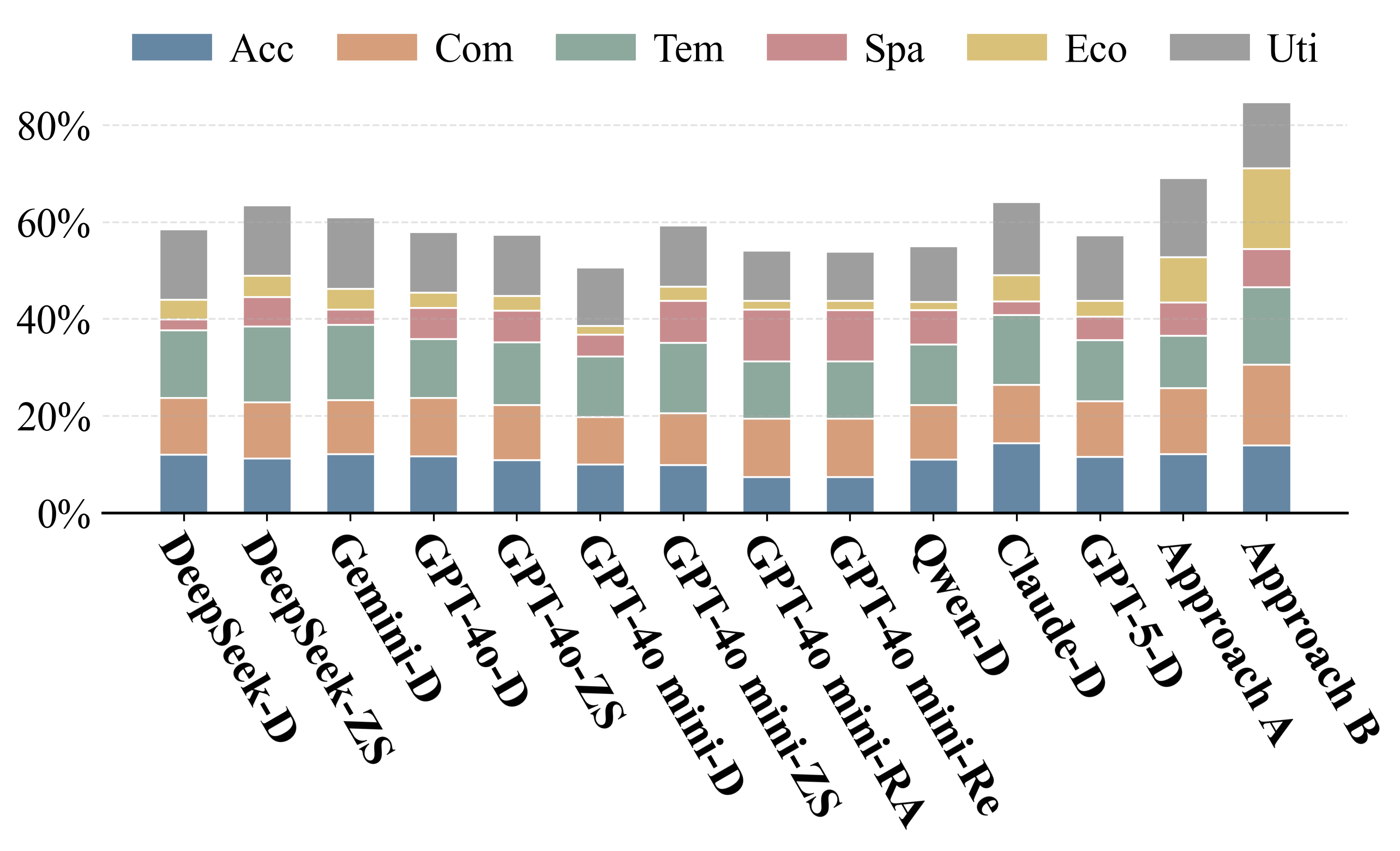}
        \caption{Easy Mode}
        \label{fig:easy}
    \end{subfigure}
    

    \begin{subfigure}[b]{0.48\textwidth}
        \centering
        \includegraphics[width=\textwidth]{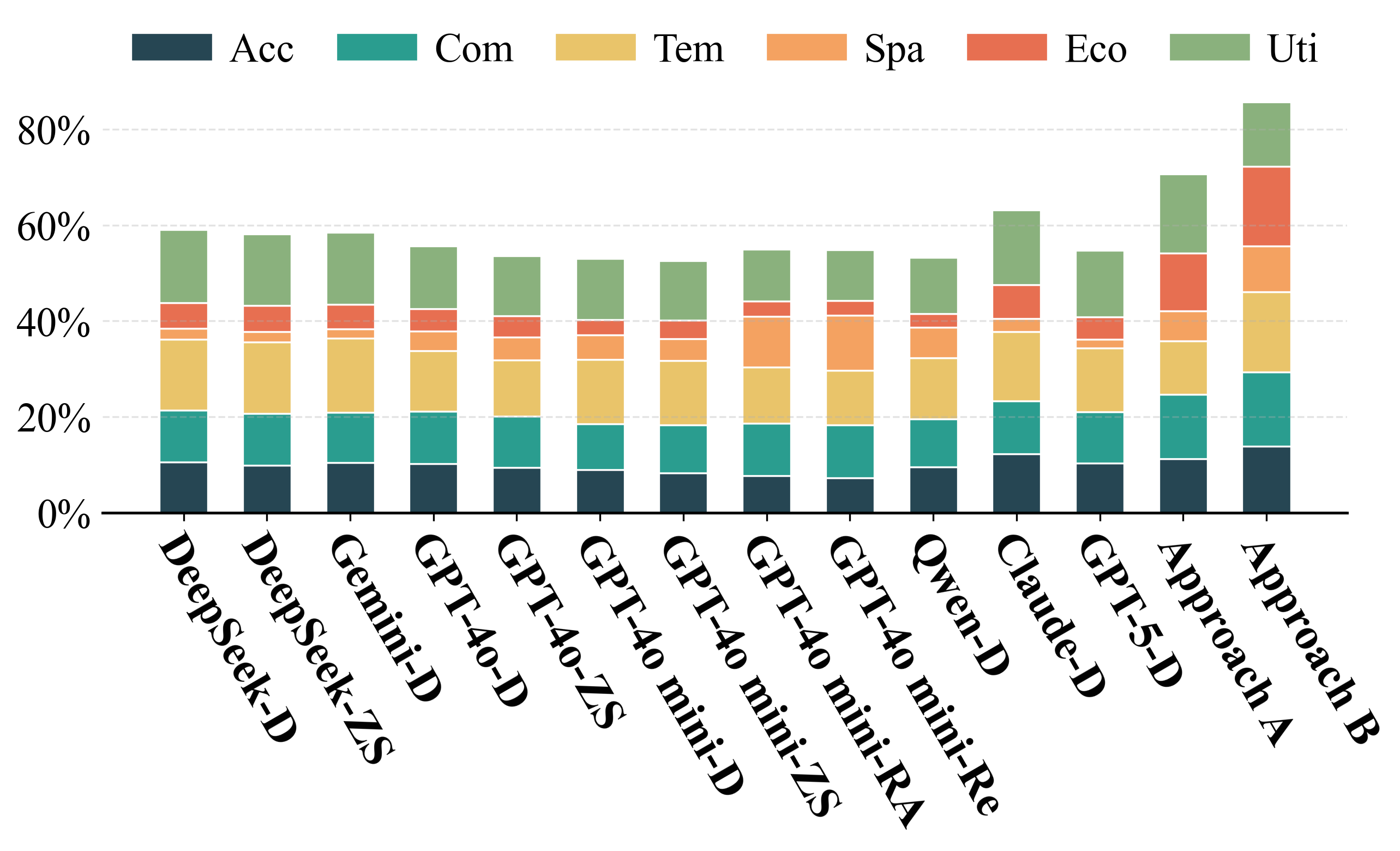}
        \caption{Medium Mode}
        \label{fig:medium}
    \end{subfigure}
    \hfill
    \begin{subfigure}[b]{0.48\textwidth}
        \centering
        \includegraphics[width=\textwidth]{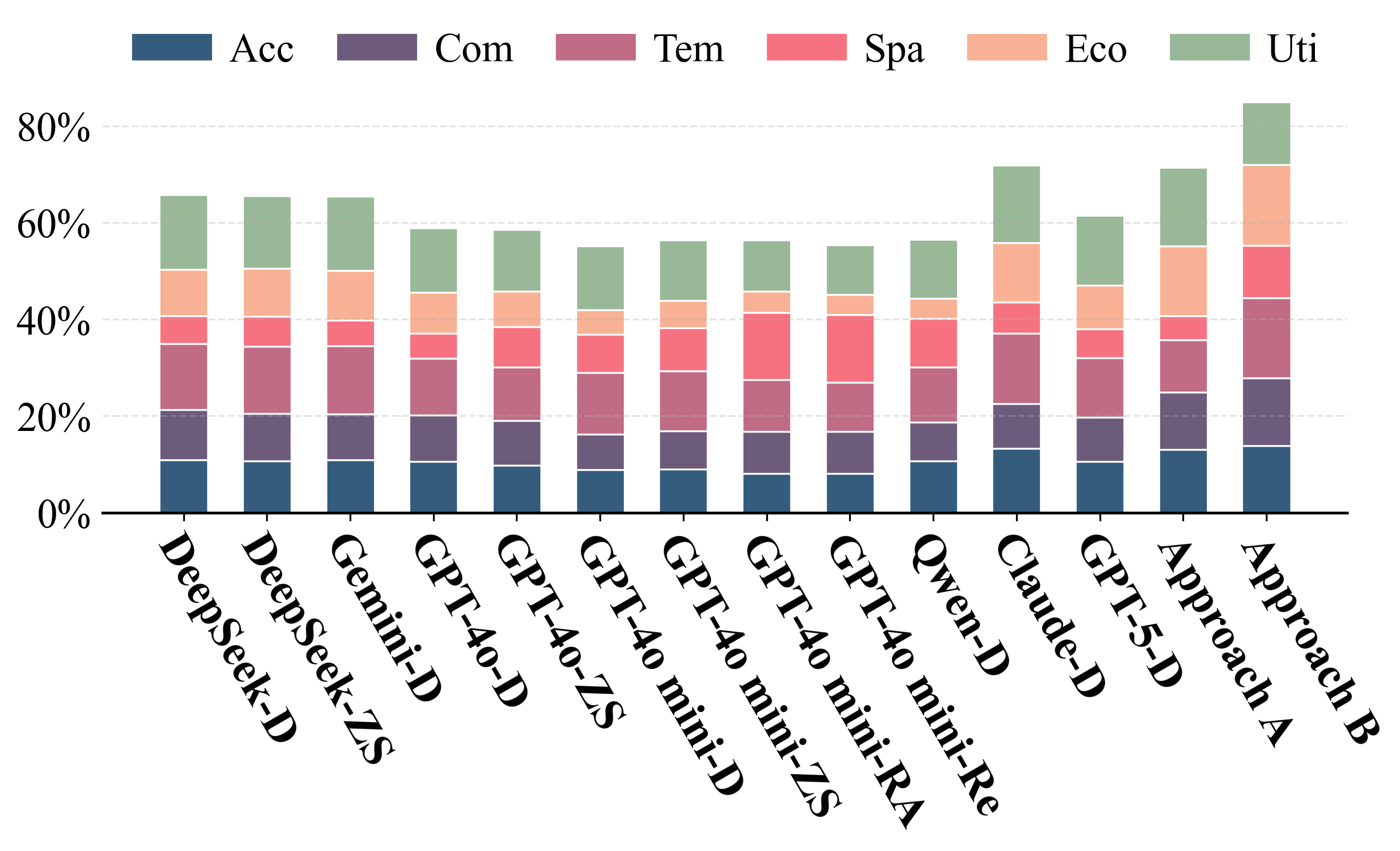}
        \caption{Hard Mode}
        \label{fig:hard}
    \end{subfigure}
    
    \caption{Comparison of six-dimension normalized scores: (a) Overall Dataset, (b) Easy Mode, (c) Medium Mode and (d) Hard Mode. For clarity, the Easy Mode, Medium Mode, and Hard Mode in this figure collectively refer to the combined results of the \textit{easy}/G1, \textit{medium}/G2, and \textit{hard}/G3 difficulty tiers, respectively.}
    \label{fig:full_comparison}
\end{figure*}

\textbf{\underline{Multi-dimensional Model Performance.}} To answer \textbf{Q1}, we summarize the LLM-generated travel planning performance based on the \texttt{TravelEval} benchmark, finding that it generally underperforms baseline approaches across key metrics (overall performance is shown in Figure~\ref{fig:overall}, with detailed breakdowns provided in Table~\ref{tab:full_results_overview}). In particular, the budget satisfaction rate consistently remains below 40\% (Figure~\ref{fig:BCS_performances}): even Claude-4.5-Sonnet, which achieves the best overall performance, only reaches 31.10\%, and GPT-4o with Direct reasoning attains 38.26\%, both substantially lower than the baseline results of 62.73\% and 92.17\%. These findings indicate that \textbf{current LLMs struggle to balance constraint compliance and planning quality}. This limitation is further illustrated by the comparison between baseline approaches (Table~\ref{tab:comparison_between_approaches}): Approach A prioritizes preference satisfaction and user experience, achieving higher in Utility, while Approach B emphasizes constraint satisfaction and practical feasibility, leading in BCS, TTCS, and the Temporality dimension. This contrast implies that \textbf{it remains difficult for current LLMs to simultaneously satisfy hard constraints, complex preferences, and user experience requirements in real-world travel planning.}

\begin{figure*}[t]
    \centering
    \includegraphics[width=\textwidth]{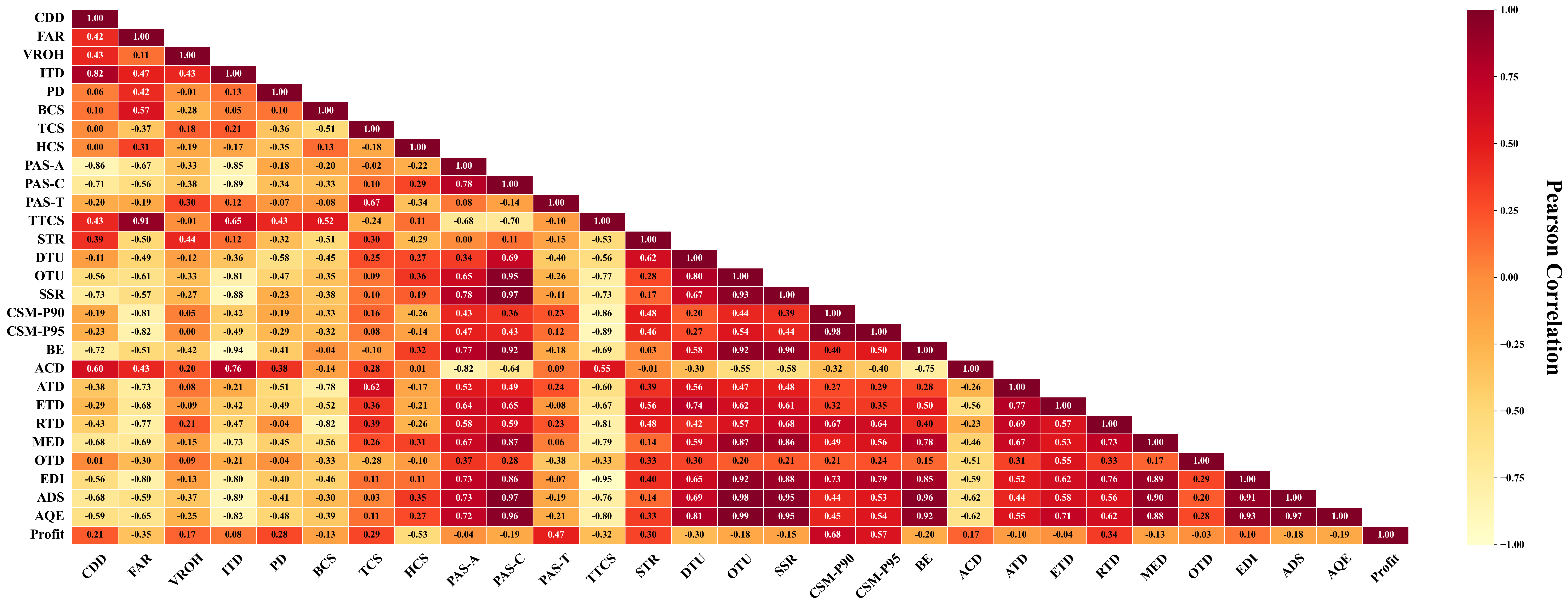}
    \caption{Pearson Correlation Coefficients between metrics.}
    \label{fig:correlation}
\end{figure*}



\textbf{\underline{Differences across Models under the Same Strategy.}} To further answer \textbf{Q1.a}, we analyze model performance under a unified strategy, with results in Figure~\ref{fig:overall}. Under the same strategy, Claude-Sonnet-4.5 performs the best, particularly in Accuracy and Utility dimensions. However, it displays notably poor performance in the TTCS and Profit, indicating that \textbf{even the state-of-the-art (SOTA) models in the current field struggle to generate feasible and reasonable travel plans while considering user preferences.} Gemini-2.0-Flash, DeepSeek-V3.1, and GPT-5-chat follow closely with balanced yet unremarkable performance.


\textbf{\underline{Differences across Strategies under the Same Model.}} To answer \textbf{Q1.b} in depth, we compare different prompting strategies for the same model and find that strategy upgrading (from Direct reasoning to Reflexion) does not significantly improve overall model capability. The performance of GPT-4o even degrades when upgrading from Direct reasoning to Zero-shot CoT (Figure~\ref{fig:overall}). Only on the CSM does the GPT series exhibit a measurable improvement with strategy upgrades. Further adding the Reflexion strategy to ReAct leads to certain improvements in TTCS and CSM, among which GPT-4o mini with Reflexion ranks first in all model-strategy combinations. While multi-round reasoning strategies (like Reflexion) enhance itinerary rationality and user-friendliness by refining attraction arrangements, \textbf{they fail to improve overall capability and notably degrade performance in Temporality and Utility dimensions compared to simpler strategies.}

\begin{table*}[htbp]
  \centering
  \caption{
  Comparison between Approach A and B.}
  \label{tab:comparison_between_approaches}
  \setlength{\tabcolsep}{4pt}    
  \renewcommand{\arraystretch}{1.3}  
  
  \begin{tabular}{>{\bfseries}c!{\vrule width 1pt}cc|ccc|cccc}
    \toprule  
    \textbf{Method} & \textbf{BCS} & \textbf{TTCS} & \textbf{STR} & \textbf{DTU} & \textbf{OTU} & \textbf{EDI} & \textbf{ADS} & \textbf{AQE} & \textbf{Profit} \\
    \midrule  
    Approach A & 62.73\% & 0.96\% & 49.79\% & 12.12\% & 6.08\% & 65.47\% & 76.49\% & 98.82\% & 5.998 \\
    Approach B & 92.17\% & 44.87\% & 76.77\% & 30.29\% & 6.49\% & 58.29\% & 51.29\% & 74.79\% & 5.502 \\
    \bottomrule  
  \end{tabular}
\end{table*}


\textbf{\underline{Differences across Task Difficulty Levels.}} To answer \textbf{Q2}, we evaluate model performance from easy to hard (Figure~\ref{fig:full_comparison}) and find that as the dataset difficulty increases (i.e., with more user preferences incorporated), several performance trends emerge. Specifically, the PAS of all models decreases significantly, while TTCS fluctuates with changes in the effective tour time of attractions. Consequently, although the Profit in Utility rises remarkably, the AQE of most model-strategy combinations shows a significant decline. Regarding ITD, affected by increasing query complexity, most models exhibit a fluctuating increase, with only Qwen achieving a noticeable reduction. This may be attributed to Qwen's access to robust transportation data via Alibaba, which is worth further research.
As difficulty increases, BE shows a trend of an initial decline followed by a rise. Compared to the Easy subset, the Hard one generally achieves nearly twice the improvement. This suggests that \textbf{LLMs demonstrate a certain adaptability when the budget gradually tightens, making more accurate and cautious budget allocations to maximize user experience within budget constraints.} However, the performance on the medium subset has declined, \textbf{indicating the model's limited adaptability. Further adjustments in this regard are needed in the future.}

\begin{table}[htbp]
\setlength{\belowcaptionskip}{2pt}
\caption{Model-strategy performance comparison across progressive difficulty levels.
$\uparrow$/$\downarrow$ indicate higher/lower is better. $\Delta$ shows relative change: $\Delta_{\text{G2}\leftarrow\text{G1}} = \frac{\text{G2}-\text{G1}}{\text{G1}}\times 100\%$, $\Delta_{\text{G3}\leftarrow\text{G2}} = \frac{\text{G3}-\text{G2}}{\text{G2}}\times 100\%$.}
\label{tab:progressive_difficulty}
\centering
\begin{tabular}{@{}llrrrrr@{}}
\toprule
\textbf{Model} & \textbf{Metric} & \textbf{G1} & \textbf{G2} & \textbf{G3} & $\boldsymbol{\Delta_{\text{G2}\leftarrow\text{G1}}}$ & $\boldsymbol{\Delta_{\text{G3}\leftarrow\text{G2}}}$ \\
\midrule
\multirow{3}{*}{\textbf{GPT5-D}}
    & FAR $\downarrow$ & 0.013 & 0.014 & 0.040 & +12.9\% & +183.9\% \\
    & OTU $\uparrow$  & 0.044 & 0.046 & 0.046 & +5.4\% & -0.0\% \\
    & AQE $\uparrow$  & 0.784 & 0.844 & 0.859 & +7.7\% & +1.8\% \\
\midrule
\multirow{3}{*}{\textbf{Qwen-D}}
    & FAR $\downarrow$ & 0.004 & 0.012 & 0.029 & +188.2\% & +134.5\% \\
    & OTU $\uparrow$  & 0.033 & 0.028 & 0.036 & -13.8\% & +27.2\% \\
    & AQE $\uparrow$  & 0.667 & 0.604 & 0.774 & -9.4\% & +28.1\% \\
\midrule
\multirow{3}{*}{\makecell{\textbf{GPT-4o-}\\\textbf{mini-RA}}}
    & FAR $\downarrow$ & 0.020 & 0.179 & 0.104 & +788.4\% & -41.7\% \\
    & OTU $\uparrow$  & 0.033 & 0.023 & 0.030 & -31.3\% & +31.9\% \\
    & AQE $\uparrow$  & 0.633 & 0.389 & 0.521 & -38.5\% & +33.8\% \\
\midrule
\multirow{3}{*}{\makecell{\textbf{GPT-4o-}\\\textbf{mini-Re}}}
    & FAR $\downarrow$ & 0.024 & 0.185 & 0.135 & +687.2\% & -27.1\% \\
    & OTU $\uparrow$  & 0.032 & 0.021 & 0.030 & -34.6\% & +41.8\% \\
    & AQE $\uparrow$  & 0.608 & 0.334 & 0.505 & -45.0\% & +51.2\% \\
\bottomrule
\end{tabular}
\end{table}

\textbf{\underline{Differences across Progressive Difficulty Settings.}} To ans\-wer \textbf{Q3}, we summarize the performance of different model-strategy combinations under the Progressive Groups and observe clear differences in overall behavior (see Table ~\ref{tab:progressive_difficulty}): GPT-5 remains the most stable with smooth cross-difficulty metric variations, Qwen shows pronounced structural instability on harder queries, while GPT-4o mini shows a rise-then-fall performance in Temporality and Utility as well as FAR metrics, versus the opposite for BCS, indicating that budget compliance sacrifices preference fulfillment and hallucination control. Notably, our analysis identifies the G2 subset as a significant bottleneck: \textbf{the need to accommodate newly added preferences increases reasoning demands for LLMs to some degree, which in turn exacerbates hallucinations and compromises attraction quality}, particularly for the ReAct and Reflexion strategies.

\textbf{\underline{Differences across LLM Generations.}} To answer \textbf{Q4}, we find that GPT-4o outperforms GPT-4o mini with a notable overall improvement, surpassing the lightweight model on all dimensions except Time. By contrast, GPT-5, the latest LLM, only yields marginal gains on specific high-correlation metrics such as ADS, AQE and EDI (correlation shown in Figure~\ref{fig:correlation}) relative to GPT-4o. Despite modest overall performance gains, it fails to attain SOTA performance on any single evaluation dimension. This suggests that while scaling up from lightweight models yields substantial gains, the latest generation exhibits diminishing returns, underscoring that \textbf{complex travel planning capabilities cannot be enhanced solely through simple model iteration.} 

\textbf{\underline{Effectiveness of External Tool Calling.}} To answer \textbf{Q5}, we evaluate the performance of the tool-augmented agent and observe a significant improvement in the BCS (see Table~\ref{tab:claude code} for partial results, and full evaluation data can be found in Table~\ref{tab:claude_code_full}). Over half of the plans meet budget constraints, effectively mitigating the foundation model's inherent cost-control shortcomings. While the agent outperforms the foundation model in Accuracy and Economy, it slightly lags across the remaining dimensions. Notably, due to a minor version discrepancy between the agent's underlying model (4.6) and the foundation model used in the main experiments (4.5), these conclusions serve as a preliminary reference, and performance on the full dataset requires further validation through comprehensive experiments. Nevertheless, results confirm that tool-calling mechanisms effectively resolve constraint conflicts in complex travel planning.

\begin{table}[htbp]
    \centering
    \caption{The results of Claude Code's experiments}
    \label{tab:claude code}
    \renewcommand{\arraystretch}{1.2} 
    \resizebox{\linewidth}{!}{
    \begin{tabular}{@{}llcccc@{}} 
        \toprule 
        \textbf{Dimension} & \textbf{Metric} & \textbf{Overall} & \textbf{Easy} & \textbf{Medium} & \textbf{Hard} \\
        \midrule 
        
        \multirow{2}{*}{\textbf{Accuracy}} 
        & CCD $\downarrow$ & 0.0407 & 0.0380 & 0.0449 & 0.0383 \\ 
        & VROH $\downarrow$ & 0.0235 & 0.0275 & 0.0273 & 0.0173 \\ \midrule 
        
        \multirow{1}{*}{\textbf{Compliance}} 
        & BCS $\uparrow$ & 0.5703 & 0.6508 & 0.8100 & 0.2800 \\ 
        \midrule
        
        \multirow{2}{*}{\textbf{Temporality}} 
        & STR $\uparrow$ & 0.6958 & 0.6900 & 0.7028 & 0.6925 \\ 
        & DTU $\uparrow$ & 0.1955 & 0.1754 & 0.1973 & 0.2064 \\ \midrule
        
        \multirow{2}{*}{\textbf{Spatiality}} 
        & SSR $\downarrow$ & 0.2830 & 0.2701 & 0.2871 & 0.2871 \\ 
        & CSM-P90 $\downarrow$ & 2.9715 & 2.4953 & 3.0774 & 3.1676 \\ \midrule
        
        \multirow{1}{*}{\textbf{Economy}} 
        & BE $\uparrow$ & 13.212 & 11.623 & 12.505 & 14.921 \\ 
        \midrule
        
        \multirow{1}{*}{\textbf{Utility}} 
        & Profit $\uparrow$ & 5.6632 & 4.9440 & 5.3539 & 6.4255 \\ 
        \bottomrule 
        
    \end{tabular}
    } 
    
\end{table}

\subsection{Experiment Insights}
\label{sec:in-depth-analysis-custom}
\begin{itemize}
    \item \textbf{Feasibility Defect in LLM-Generated Planning.} Compared to baseline approaches, LLM-generated planning exhibits a severe deficiency in feasibility, with particularly poor performance on budget satisfaction. We believe this gap to be due to the probabilistic, autoregressive decoding paradigm of LLMs and the absence of tool use, which impairs their ability to retrieve granular travel information and execute precise numerical reasoning for budget adherence. This warrants in-depth follow-up research to explore methods that can enforce strict boundary conditions without sacrificing the richness of the generated plans.
    \item \textbf{Dimensional Imbalance of LLM Capabilities.} Under an identical prompting strategy, different LLMs tend to perform better on different evaluation dimensions. This pattern suggests that the advantages of current models are often localized to certain aspects (e.g., content accuracy or experiential utility), while feas\-ibility-critical components are still a common bottleneck that is not reliably resolved by simply switching to stronger backbones.
    \item \textbf{Intrinsic Model Capability Bottleneck.} Compared to basic Direct reasoning, upgrading to more complex prompting strategies does not yield significant overall improvements. We contend that this indicates the primary bottleneck lies in the intrinsic capabilities of LLMs themselves rather than in the reasoning framework. Thus, future research should shift its focus from elaborating prompt engineering techniques to investigating and enhancing the fundamental planning proficiency of the models.
    \item \textbf{Gradual Adaptation Failure.} When confronted with the gradient increase of user preference quantities and budget constraints, LLMs' adaptive capacity fails in the Medium, only performing effectively in extreme scenarios with low or high constraints. This reflects their lack of progressive refined reasoning and resource allocation capabilities for combined constraints of multiple preferences and moderate budgets.
    \item \textbf{Cognitive Overload in Reasoning Strategies.} In contrast to GPT-5's robust stability, the non-linear collapse of other model-strategy combinations at G2 exposes a key limitation of complex reasoning strategies. We hold this reflects a critical~\textquotedbl cognitive overload\textquotedbl~threshold: as constraints accumulate, verbose intermediate reasoning clutters the context, turning reasoning aids into noise and triggering hallucinations. This indicates limited scalability in current paradigms and motivates adaptive mechanisms that manage constrained planning complexity without overwhelming the model.
    \item \textbf{Scaling Mismatch.} We expected larger models to bring qualitative leaps, yet our findings confirm generic scaling does not stably improve travel planning and even leads to regression in advanced models. This indicates a fundamental misalignment between model scaling and the unique demands of multi-constraint travel planning, warranting future research into dedicated mechanisms that enhance long-horizon planning and global constraint integration beyond simple model iteration.
\end{itemize}

\section{Conclusion}
\label{sec:conclusion}

In this paper, we propose \texttt{TravelEval}, a realistic and comprehensive benchmark for evaluating travel planning agents. It provides a six-dimensional evaluation framework with a
realistic data sandbox and a simulation-based global evaluation method. Evaluations on 12 mainstream approaches reveal even SOTA LLMs struggle with multi-dimensional global planning, thus positioning \texttt{TravelEval} as a robust foundation for advancing LLM-powered travel applications. \texttt{TravelEval} further offers clear, empirically grounded guidance for future research, identifying key technical bottlenecks in multi-constraint optimization and preference aggregation to inform the design of more capable and practical travel planning agents.


\begin{acks}
\sloppy
This work is partially supported by the ``Pioneer'' and ``Leading Goose'' R\&D Program of Zhejiang (No. 2026C02A1236) and the Open Research Fund of Zhejiang Key Laboratory of Intelligent Education Technology and Application (No. 2025ZNJYKF011). Additional support is provided by the NSFC/RGC Joint Research Scheme (N\_PolyU5179/25), RGC Hong Kong (PolyU25600624), Innovation Technology Fund (ITS/052/23MX, PRP/009/22FX), and the RCDTT Project (P0058790, RAG4Tourism: Advancing Transparent and Collaborative Travel Planning with LLM). Further support comes from National Key R\&D Program of China (No. 2023YFF0725100), National Science Foundation of China (NSFC) (No. U22B2060), Guangdong-Hong Kong Technology Innovation Joint Funding Scheme (No. 2024A0505040012), the Hong Kong RGC GRF Project (16213620), RIF Project (R6020-19), AOE Project (AoE/E-603/18), Theme-based project TRS (T41-603/20R), CRF Project (C2004-21G), Key Areas Special Project of Guangdong Provincial Universities (2024ZDZX1006), Guangdong Province Science and Technology Plan Project (2023A0505030011), Guangzhou Municipality Big Data Intelligence Key Lab (2023A03J0012), Hong Kong ITC (TC-SKLCRCC26EG01), Hong Kong ITC ITF grants (MHX/078/21, PRP/004/22FX), Zhujiang Scholar Program (2021JC02X170), Microsoft Research Asia Collaborative Research Grant, HKUST-Webank Joint Research Lab, 2025 HKUST Shenzhen-Hong Kong Collaborative Innovation Institute Green Sustainability Special Fund from Shui On Xintiandi and InnoSpace GBA, and HKUST (GZ) - CMCC (Guangzhou Branch) Metaverse Joint Innovation Lab (No. P00659). We also acknowledge the National Science Foundation of China (NSFC) (No. 62506075).

\end{acks}

\bibliographystyle{ACM-Reference-Format}
\balance
\bibliography{ref}

\clearpage
\appendix

\counterwithin{table}{section}
\counterwithin{figure}{section}
\renewcommand{\thetable}{\Alph{section}\arabic{table}}
\renewcommand{\thefigure}{\Alph{section}\arabic{figure}}

\section{Additional Information}
\label{sec:Appendix_A}
In this section, we provide the additional information of the research process.
\subsection{Metric Details}
\label{sec:Appendix_A1}
In this subsection, we detail each evaluation metric as
follows:

\textbf{Accuracy Dimension.} Evaluate the authenticity of information and the reliability of calculations.
    \begin{itemize}
        \item \textbf{Cost Calculation Deviation (CCD).} Quantitative arithmetic deviation of LLM in cost calculation.
        The CCD is defined as:
       \begin{equation}
            \text{CCD}(A_{d,i}) = \frac{\sum_{i=1}^n \left| C_{AI_i} - C_{actual_i} \right|}{C_{actual}}
        \end{equation}
        where $C_{AI_i}$ and $C_{actual_i}$ represents the $i\text{-th}$ cost calculated by LLM and the actual cost, while $C_{actual}$ denotes the total actual cost. The deviation should be computed for each individual item.
        
        \item \textbf{Fictitious Attraction Rate (FAR).} Detect whether LLM has fabricated non-existent tourist attractions. 
        The FAR is defined as: 
        \begin{equation}
            \text{FAR}= \frac{\sum_{j=1}^{k} \mathbb{I}\big(v_j = 0\big)}{k}
        \end{equation}
        where $\mathbb{I}(v_j)$ denotes the indicator function, where $v_j=1$ if attraction $j$ passes the verification, and $v_j=0$ otherwise. The symbol $k$ represents the total number of recommended attractions.

        \item \textbf{Violation Rate of Opening Hours (VROH).} Count the arrangements that violate the actual opening hours of attractions (only counted towards arrangements that are completely within the closed hours). The VROH is defined as:
        \begin{equation}
            \text{VROH}= \sum_{d=1}^{D} \sum_{j=1}^{A_d} \mathbb{I}\big( [s_{dj}, e_{dj}] \cap O_j = \emptyset \big)
        \end{equation}
        where $D$ represents the total number of travel days, $A_d$ denotes the number of attractions scheduled on day $d$,
        $[s_{dj}, e_{dj}]$ stands for the planned visiting time interval of attraction $j$ on day $d$,
        and $O_j$ indicates official opening time interval of attraction $j$.

        \item \textbf{Failure Rate of Traffic Connection (FRTC).} Count the arrangements that violate the actual opening hours of the scenic spots (only detecting the arrangements that are completely within the closed hours). The FRTC is defined as:
        \begin{equation}
            \text{FRTC}= \sum_{r=1}^{R} \mathbb{I}\big(\phi_r = 1\big)
        \end{equation}
        where $R$ represents the total number of transportation segments, and $\phi_r$ denotes the feasibility status of transportation segment $r$ ($\phi_r = 1$ for infeasible, $\phi_r = 0$ for feasible). Specifically, $\phi_r$ is set to 1 if the official API verifies no trips on the current day, or if the last trip time is earlier than the planned departure time; otherwise, $\phi_r$ is 0.

        \item \textbf{Intercity Traffic Deviation (ITD).} Detecting infeasible transportation connections. The ITD is defined as:
        \begin{equation}
            \text{ITD} = \frac{|\,T_{plan} - T_{actual}\,|}{T_{actual}}
        \end{equation}
        where $T_{\text{plan}}$ denotes the pre-planned travel time for intercity trips, and $T_{\text{actual}}$ denotes the actual travel time consumed for intercity trips ($\text{ITD} = 0$ means the planned time is exactly consistent with the actual time, and the larger the value of $\text{ITD}$, the more serious the deviation between the planned time and the actual time).

        \item \textbf{Personnel Deviation (PD):} Evaluate whether the travel plan is fully in line with the number of tourists.
        \begin{equation}
            \text{PD} = \frac{\left|n_{{bed}} - n\right| + \left|n_{{veh}} \cdot l_{{load}} - n\right| + \left|q_{{tkt}} - n\right|}{n}
        \end{equation}
        where $n$ denotes the total number of tourists, $c_{{bed}}$ represents the available accommodation capacity, $n_{veh}$ and $L_{{load}}$ denote the vehicle count and the per-vehicle load limit respectively (determining the transport capacity), and $q_{{tkt}}$ represents the attraction ticket quota.
    \end{itemize}

\textbf{Compliance Dimension.} Evaluate to what extent the LLM travel planning system can meet the specific demands and preferences of users, and ensure that the planning results align with the basic expectations of the users.
    \begin{itemize}
        \item \textbf{Budget Compliance Sign (BCS):} It takes the value of 1 if the total cost is within the user's budget, and 0 otherwise.

        \item \textbf{Time Compliance Sign (TCS):} It takes the value of 1 if the number of planned days matches the number of days required by the user, and 0 otherwise.

        \item \textbf{Headcount Compliance Sign (HCS):} Evaluate whether the travel plan is suitable for the number of tourists involved in the trip. The HCS is defined as:
        \begin{equation}
            {Adapt}_i=
            \begin{cases}
            1 & \text{if } {Supply}_i \geq N_{{user}} \\
            0 & \text{otherwise}
            \end{cases}
        \end{equation}

        \begin{equation}
            \text{HCS} = \prod_{i \in \{{acc}, {trans}, {tick}\}} {Adapt}_i
        \end{equation}
        where $i \in \{{acc},~{trans},~{tick}\}$ represents accommodation, transportation, and ticket scenarios respectively; \({Adapt}_i\) denotes the number adaptability of the \(i\text{-th}\) scenario, which takes the value of 1 if the planned supply quantity \({Supply}_i\) is greater than or equal to the total number of users \(N_{{user}}\), and 0 otherwise. 

        \item \textbf{Preference Achievement Sign (PAS):} It takes the value of 1 if the planned accommodation type (PAS-A), the planned transportation mode (PAS-T), and the planned cuisine type (PAS-C) match the user preferred ones, and 0 otherwise.

        \item \textbf{Tour-Time Compliance Sign (TTCS):} Check whether actual attraction tour time is within the recommended reasonable range. The TTCS is defined as:
        \begin{equation}
        \text{TTCS} = 
        \begin{cases}
            1 & \text{if } R_{i-\min} \leq T_i \leq R_{i-\max}, \\
            0 & \text{otherwise}.
        \end{cases}
        \end{equation}
        where $R_{\text{min},i}$ and $R_{\text{max},i}$ denote the sandbox-recommended minimum and maximum visit duration for attraction $i$,
        and $T_i$ indicates the planned visit duration allocated to attraction $i$ in the travel plan.
    \end{itemize}

\textbf{Temporality Dimension.} Evaluate the efficiency and rationality of time utilization in travel planning.
\begin{itemize}
    \item \textbf{Scene-Time Ratio (STR):} Ratio of effective tour time (considering queuing time) to total allocated time per attraction. The STR is defined as:
    \begin{equation}
        \text{STR}=\begin{cases}
        \dfrac{E_i}{P_{i-{total}}}, & R_{i-\min} \leq E_i \\
        0, & E_i < R_{i-\min}
        \end{cases}
    \end{equation}
    where $R_{{min},i}$ represents the sandbox-recommended minimum visit duration for attraction $i$, while $P_{\text{total},i}$ indicates the total time allocated to attraction $i$ in the travel plan, including queuing time $Q_i$ and actual visit duration $E_i$ ($E_i = P_{\text{total},i} - Q_i$).
    
    \item \textbf{Daily Time Utilization (DTU):} Evaluate the proportion of effective visit duration in daily travel time. The DTU:
    \begin{equation}
        \text{DTU}= \frac{\sum_{i=1}^{n} E_i}{T_{\text{total}}}
    \end{equation}
    where $i \in \{1, 2, ..., n\}$ represents the $i\text{-th}$ attraction in the itinerary for the day, $E_i$ denotes the effective visit duration of attraction $i$, and $T_{\text{total}}$ denotes the total time from the first departure (e.g., a hotel or station) to the final return (e.g., a hotel or station) on the day, covering all time elements in the travel plan (including attraction time, transportation time, dining time, rest time, etc.).

    \item \textbf{Overall Time Utilization (OTU):} Evaluate the proportion of effective attraction visit duration in the total travel planning. The OTU is defined as:
    \begin{equation}
        \text{OTU}=\frac{\sum E_i'}{T_{\text{total}}} \times 100\%
    \end{equation}
    where $\sum_{i} E_i'$ represents the total effective visiting time for all attractions $i$ in the travel plan,
      and $T_{\text{total}}$ denotes the total travel time of the entire itinerary.
    
\end{itemize}
        
\textbf{Spatiality Dimension.} Evaluate the rationality of the distribution of attractions and the route in the travel plan.
\begin{itemize}
    \item \textbf{Scene Sequence Rationality (SSR):} Quantitatively evaluate the rationality of the visit sequence of scenic spots and penalize unreasonable routes.This metric is measured by the Route Penalty. The Route Penalty is defined as: 
    \begin{equation}
        \text{RP} = \frac{L_{actual}}{L_{optimal}} -1
    \end{equation}
    where $L_{\text{actual}}$ is the total driving distance of the generated route retrieved via the API, and $L_{\text{optimal}}$ denotes the total distance of the theoretical shortest path. The RP metric reflects the degree of deviation from the optimal trajectory, implying that a lower value signifies a shorter travel distance and higher route efficiency.

    \item \textbf{Cross-Day Spatial Mismatch (CSM):} For a given attraction \(A_{d,i}\) on day \(d\), the CSM is calculated by comparing: \textit{(1)} The average distance between \(A_{d,i}\) and other attractions on the same day (\(\text{AvgDist}_{d}\)). \textit{(2)} The minimum average distance between \(A_{d,i}\) and attraction clusters on other days (\(\min_{k \neq d} \text{AvgDist}_{k}(i)\)).
   
    The CSM is defined as:
    \begin{equation}
        \text{CSM}(A_{d,i}) = \frac{\text{AvgDist}_{\text{d}}(A_{d,i}, Cluster_{d})}{\min_{k \neq d} \text{AvgDist}_{\text{k}}(A_{d,i}, \text{Cluster}_k)}
    \end{equation}
where $A_{d,i}$ represents the attraction $i$ scheduled on day $d$, and $\text{AvgDist}_d(\cdot)$ denotes the average distance from $A_{d,i}$ to all other spots on the same day. The denominator represents the minimum average distance from $A_{d,i}$ to the clusters of any other day $k$ (where $k \neq d$). 
\end{itemize}

\textbf{Economy Dimension.} Evaluate the proportion of expenses for each part in the tourism process
\begin{itemize}
    \item \textbf{Budget Efficiency (BE):} Ratio of total experience value to actual expenditure, quantifying the model's budget conversion capability. The BE is defined as:
    \begin{equation}
        \text{BE} = \frac{\sum_{i=1}^{n} S_i \cdot M(T_i)}{U} \times 100\%
    \end{equation}
    where $n$ denotes the total number of attractions in the travel plan,
      $S_i$ indicates the user satisfaction score of attraction $i$,
      $M(T_i)$ represents the time matching degree function of attraction $i$, which measures the rationality of the allocated visiting time $T_i$,
      and $U$ denotes the actual total cost of the LLM -planned travel scheme.
    \item \textbf{Cost Composition:} This metric decomposes the total travel expenditure into six cost categories to characterize the budget allocation structure. 
    \begin{itemize}
        \item \textbf{Accommodation Cost Distribution (ACD):} The proportion of total cost spent on accommodations, including hotels and other lodging-related expenses.
        \item \textbf{Attractions Cost Distribution (ATD):} The proportion of total cost allocated to attraction entrance fees and on-site activity expenses.
        \item \textbf{Intercity Transportation Cost Distribution (ETD):} The proportion of total cost incurred by intercity transportation, such as flights and high-speed rail between cities.
        \item \textbf{Intracity Transportation Cost Distribution (RTD):} The proportion of total cost spent on intracity transportation, including metro, bus, taxi, and other local travel within a city.
        \item \textbf{Meals Cost Distribution (MED):} The proportion of total cost allocated to food and dining expenses.
        \item \textbf{Other Cost Distribution (OTC):} The proportion of remaining miscellaneous costs that do not fall into the above categories.
    \end{itemize}
    Together, these components reflect how the model distributes budget across essential and discretionary aspects of the travel plan, providing a fine-grained view of budget structure rationality.
\end{itemize}

\textbf{Utility Dimension.} Evaluate the value of the travel planning experience for tourists.
\begin{itemize}
    \item \textbf{Experience Diversity Index (EDI)} 
    \begin{equation}
        \text{S}_i = \sum_{j \in \text{type } i} \frac{1}{n_j}
    \end{equation}
    where the score increment for each experience type in attraction $j$ is $\frac{1}{n_j}$，

     For each attraction $a$ in the plan, if it contains $n_a$ experience types, the contribution of this attraction to a specific type $i$ is defined as:
    \begin{equation}
        c_{a,i} = \begin{cases} 
            \frac{1}{n_a} & \text{if attraction } a \text{ includes type } i \\
            0 & \text{otherwise}
        \end{cases}
    \end{equation}

    \item \textbf{Aggregation of Type Scores:} 
    The total score $S_i$ for experience type $i$ is obtained by summing the contributions of all attractions in the plan $\mathcal{A}$:
    \begin{equation}
        S_i = \sum_{a \in \mathcal{A}} c_{a,i}
    \end{equation}

    \item \textbf{Probability Calculation:} 
    The probability $p_i$ for each type $i$ is calculated as the ratio of $S_i$ to the sum of scores for all types:
    \begin{equation}
        p_i = \frac{S_i}{\sum_{j=1}^{N} S_j}
    \end{equation}

    \item \textbf{Shannon Entropy Calculation:} 
    Substitute $p_i$ into the Shannon entropy formula. For types not present in the plan ($p_i = 0$), the term $p_i \log_2(p_i)$ is treated as 0:
    \begin{equation}
        H = -\sum_{i=1}^{N} p_i \log_2(p_i)
    \end{equation}

    \item \textbf{Normalization (EDI):} 
    Finally, the entropy is normalized by the maximum possible entropy ($\log_2 N$) to obtain the Experience Diversity Index (EDI) in the range $[0, 1]$, where $N$ is the total number of pre-defined types:
    \begin{equation}
        EDI = \frac{H}{\log_2(N)}
    \end{equation}

    \item \textbf{Attractions Density Score (ADS):} Evaluate the alignment between daily attraction quantity and user expectations.
    The ADS is defined as:
    \begin{equation}
        \text{ADS} = \frac{Att_{act}}{Att_{tar}}
    \end{equation}
    where $Att_{act}$ denotes the actual number of attractions arranged in the LLM-planned itinerary for one day,
      and $Att_{tar}$ indicates the target number of attractions per day, whose value is defined according to different tourist group types (such as special forces-style tourists, solo travelers, parent-child travelers, elderly travelers, etc.).

    \item \textbf{Attractions Quality Efficiency (AQE):} Assess if the system prioritizes high-quality (high-score) attractions, reflecting the efficiency of time allocation relative to experience quality. 
    The calculation formula is:
    \begin{equation}
        \text{AQE} = \frac{\sum(S_i \cdot e_i)}{D_{\text{sum}}}
    \end{equation}    
    
where  $S_i$ is the objective score of attraction $i$, $e_i$ represents the effective tour time at attraction $i$, and $D_{\text{sum}}$ denotes the total tour time.

    \item \textbf{Profit:} The overall experience value of the trip is derived from the weighted sum of each attraction's objective score \( S_i \) and its preference match function \( M(T_i) \). If types of an attraction align with user preferences, the corresponding match function is set to \( 1 + | \text{user preference intensity} | \). The specific rules are as follows:

\begin{enumerate}
    \item \textbf{Compatibility with ``Undeclared Preferences'':} If the user has no clear preference, \( M(T_i) = 1 \) (which simplifies to the average objective score).
    \item \textbf{Support for Multiple Preference Overlap:} If the user simultaneously prefers multiple types (e.g., historical sites and Instagram-Worthy Locations), \( M(T_i) \) is incremented by 1 for each matched type.
    \item \textbf{One-Vote Veto for Rejected Types:} If the user explicitly rejects a certain type (e.g., museums), then \( M(T_i) = 0 \).
\end{enumerate}

The formula is:

\begin{equation}
    \text{Profit} = \frac{\sum (S_i \times M(T_i))}{n}
\end{equation}

where \( n \) is the total number of attractions, \( S_i \) is the objective score of attraction \( i \), \( T_i \) is the type set of attraction \( i \), and \( M(T_i) \) is the preference matching function. Detailed calculation rules for \( M(T_i) \).
\end{itemize}

\subsection{Sandbox and Evaluation Method Details}
\label{sec:Appendix_A2}
In this subsection, we detail some relevant parameters.

\textbf{Sandbox}. Our dataset encompasses 3,411 attractions, 4,976 accommodations, 4,655 restaurants, 1,703 flights, 8,089 trains and 19,584 POIs, with detailed attribute information for each category shown in Table~\ref{tab:poi-info}. The user query dataset totals 1,150 entries, consisting of 200 easy, 400 medium, 400 hard, and 150 progressive difficulty queries (Table~\ref{tab:queries_detail}).

\begin{table}[htbp]
\centering
\setlength{\abovecaptionskip}{2pt}
\setlength{\belowcaptionskip}{2pt}
\vspace{-2pt}
\caption{Dataset Distribution.}
\label{tab:queries_detail}
\setlength{\tabcolsep}{5pt}
\renewcommand{\arraystretch}{0.92}
\begin{tabular*}{\columnwidth}{@{\extracolsep{\fill}} l c c c c c c @{}}
\toprule
\textbf{queries} & \textbf{2-day} & \textbf{3-day} & \textbf{4-day} & \textbf{5-day} & \textbf{6-day} & \textbf{7-day} \\
\midrule
Easy        & 20 & 30 & 40 & 40 & 30 & 40 \\
Medium      & 40 & 60 & 80 & 80 & 60 & 80 \\
Hard        & 40 & 60 & 80 & 80 & 60 & 80 \\
Progressive & 15 & 24 & 30 & 30 & 24 & 27 \\
\bottomrule
\end{tabular*}
\vspace{-2pt}

\end{table}

\textbf{Evaluation Methods}. We established an empirical queue time prediction model, with category, date, and time-slot coefficients calibrated using user-contributed data (Table~\ref{tab:queue_time_params}). For the attraction density baseline, we derived reference values of daily attraction counts for four travel types (slow travel, family tour, regular tour, and special forces travel) from travel logs, providing both mean and median benchmarks (Table~\ref{tab:attraction_stats}). Additionally, we have completed a one-factor-at-a-time (OFAT) sensitivity analysis as sho\-wn in Table~\ref{tab:sensitivity_analysis}. Results show that K\_date is the dominant factor causing queuing fluctuations, while other parameters also exert certain influences on queuing time. This finding can guide LLMs to rearrange attraction orders and avoid peak hours. To alleviate small-sample bias, we use the median for calibration based on public travel logs. Since the original PDF is unmodifiable, we will expand the sample size to further improve the model in future work.

\begin{table}[htbp]
    \centering
    \caption{Parameters of attraction density baseline.}
    \label{tab:attraction_stats}
    \renewcommand{\arraystretch}{1.1}
    \begin{tabular*}{\columnwidth}{@{\extracolsep{\fill}} l c c @{}}
        \toprule
        \textbf{Travel Type} & \textbf{Mean} & \textbf{Median} \\
        \midrule
        Slow Travel                  & 2.24 & 2.00 \\
        Family Tour                  & 2.82 & 2.75 \\
        Regular Tour                 & 3.67 & 3.54 \\
        Special Forces Travel  & 5.56 & 6.00 \\
        \bottomrule
    \end{tabular*}
\end{table}

\begin{table}[htbp]
\setlength{\abovecaptionskip}{2pt}
\setlength{\belowcaptionskip}{2pt}
\vspace{-4pt}
    \caption{Sensitivity Analysis Ranges of Queuing Time Model}
    \small
    \label{tab:sensitivity_analysis}
    \renewcommand{\arraystretch}{1.3}
    \begin{tabular*}{\columnwidth}{@{\extracolsep{\fill}} l c c c r @{}}
        \toprule
        \textbf{Parameter} & \textbf{Minimum} & \textbf{Maximum} & \textbf{Range} & \textbf{Range/Baseline} \\
        \midrule
        $B_{category}$ & 3.3 & 11     & 7.7    & 233.33\% \\
        \addlinespace 
        $K_{date}$     & 3.3 & 33.198 & 29.898 & 906.00\% \\
        \addlinespace
        $K_{time}$     & 2.1 & 3.3    & 1.2    & 36.36\%  \\
        \bottomrule
    \end{tabular*}
    \vspace{-4pt}

\end{table}


\begin{table*}[htbp]
  \centering
  \caption{POI Information}
  \label{tab:poi-info}
  \begin{tabularx}{\textwidth}{llX}
    \toprule
    \textbf{POI} & \textbf{Counts} & \textbf{Information} \\
    \midrule
    Attractions & 3411 & Id, Name, Type, Price, Latitude, Longitude, Opentime, Endtime, Recommendmintime, Recommendmaxtime, Star \\
    \addlinespace
    Accommodations & 4976 & Id, Name, Hotelname\_en, Featurehoteltype, Latitude, Lontitude, Single\_room\_price, Single\_room\_stock, King\_room\_price, King\_room\_stock, Double\_bed\_price, Double\_bed\_stock, Family\_room\_price, Family\_room\_stock\\
    \addlinespace
    Restaurants & 4655 & Id, Name, Latitude, Lontitude, Price, Cuisine, Opentime, Endtime, Recommendedfood \\
    \addlinespace
    Flights & 1703 & FlightID, From, To, BeginTime, EndTime, Duration, Cost \\
    \addlinespace
    Trains & 8089 & TrainID, TrainType, From, To, BeginTime, EndTime, Duration, Cost \\
    \addlinespace
    POIs & 19584 & Name, Position (Latitude, Longitude)\\
    \bottomrule
  \end{tabularx}
\end{table*}

\begin{table*}[htbp]
    \centering
    \begin{threeparttable}
        \caption{Parameters for Estimated Queue Time Calculation ($T_{est}$)}
        \label{tab:queue_time_params}
        
        \begin{tabularx}{\textwidth}{Xcc}
            \toprule
            \textbf{Item} & \textbf{Value} & \textbf{Notes} \\
            \midrule
            
            Cultural Landscape  & 3.0 & Median time (min), 4 weekday samples \\
            Gardens             & 3.0 & Median time (min), 4 weekday samples \\
            Natural Scenery     & 3.5 & Median time (min), 2 weekday samples \\
            Historical Sites    & 4.0 & Median time (min), 4 weekday samples \\
            Amusement Parks     & 6.0 & Median time (min), 5 weekday samples \\
            Museums             & 10.0 & Median time (min), 6 weekday samples \\
            
            \midrule
            
            Grade S - Golden Week   & 10.06 & $\approx 10\times$ weekday baseline \\
            Grade B - Weekend       & 7.55  & $\approx 7.5\times$ weekday baseline \\
            Grade A - Special Peak  & 4.63  & $\approx 4.6\times$ weekday baseline \\
            Grade A - Short Holiday & 4.39  & $\approx 4.4\times$ weekday baseline \\
            Grade C - Winter/Summer & 3.94  & $\approx 3.9\times$ weekday baseline \\
            Grade D - Weekday       & 1.00  & Baseline \\
            
            \midrule
            
            Opening (08:00--10:00)      & 0.85 & $-15\%$ relative to average \\
            Late Morning (10:00--12:00) & 0.90 & $-10\%$ relative to average \\
            Afternoon (12:00--16:00)    & 1.10 & $+10\%$ relative to average \\
            Evening (16:00--20:00)      & 1.10 & $+10\%$ relative to average \\
            Closing (20:00--Close)      & 0.70 & $-30\%$ relative to average \\
            
            \bottomrule
        \end{tabularx}
        
        \begin{tablenotes}
            \small
            \item \textit{Note:} The estimated queue time is calculated as $T_{queue} = B_{category} \times K_{date} \times K_{time}$.
        \end{tablenotes}
    \end{threeparttable}
\end{table*}

\subsection{Evaluated Models and Settings}
\label{sec:Appendix_A3}

We benchmark 7 representative LLM backbones (covering proprietary and open-source families, as shown in Table~\ref{tab:model_details}) under 4 reasoning strategies to isolate the effects of model capacity and agentic prompting. We select these models to provide a balanced coverage of (i) strong proprietary systems commonly adopted in real-world assistants, (ii) fast or lightweight variants that reflect practical cost-latency constraints, and (iii) competitive open-source backbones that facilitate reproducibility and broader accessibility. This suite also spans different provider ecosystems and training lineages, which helps reveal whether travel-planning weaknesses are model-specific or systematic under shared prompts and evaluation.

\begin{figure*}[t] 
    \centering
    \begin{subfigure}[b]{0.48\textwidth}
        \centering
        \includegraphics[width=\textwidth]{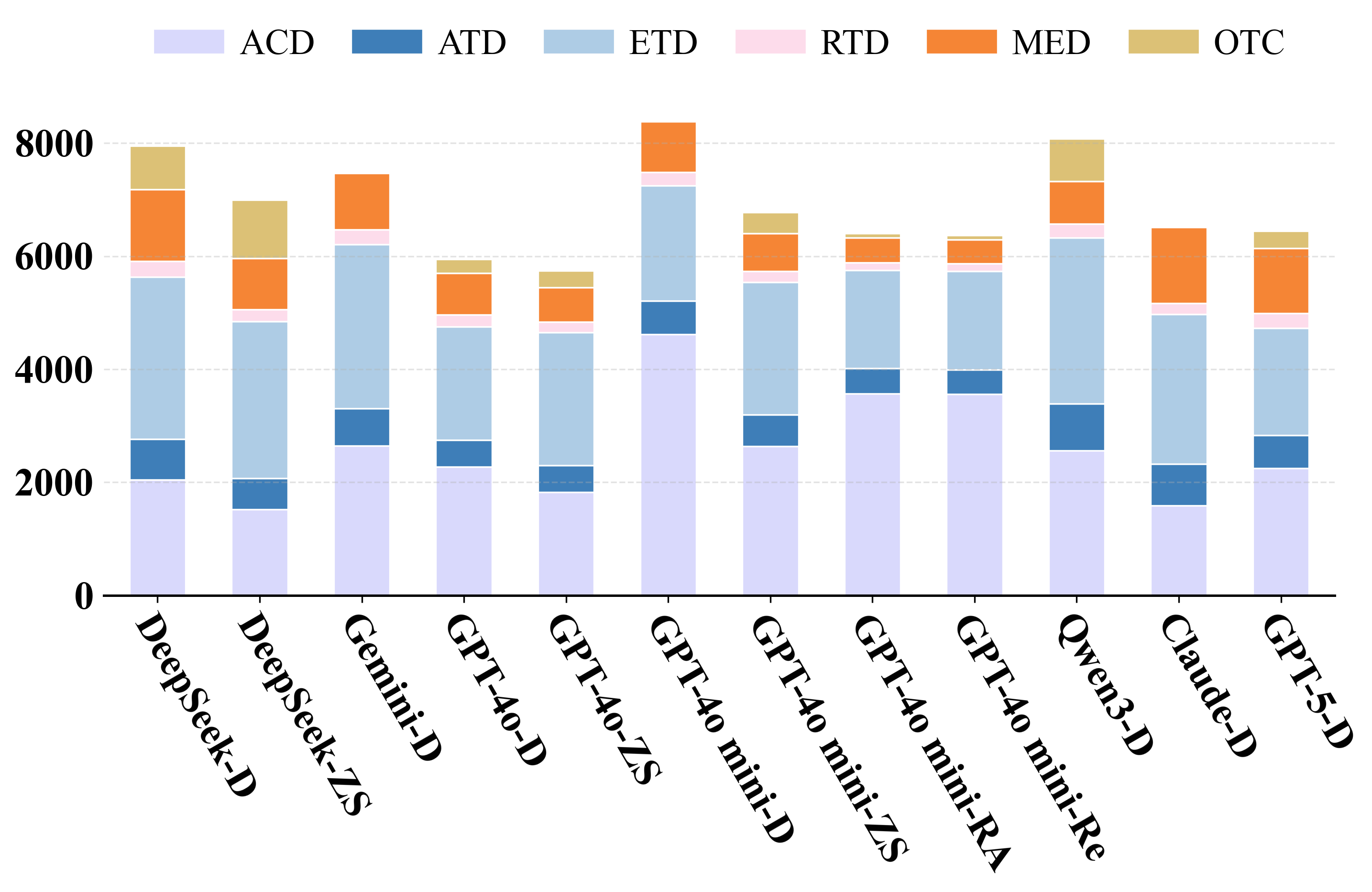}
        \caption{Overall Dataset} 
        \label{fig:overall_Cost}
    \end{subfigure}
    \hfill 
    \begin{subfigure}[b]{0.48\textwidth}
        \centering
        \includegraphics[width=\textwidth]{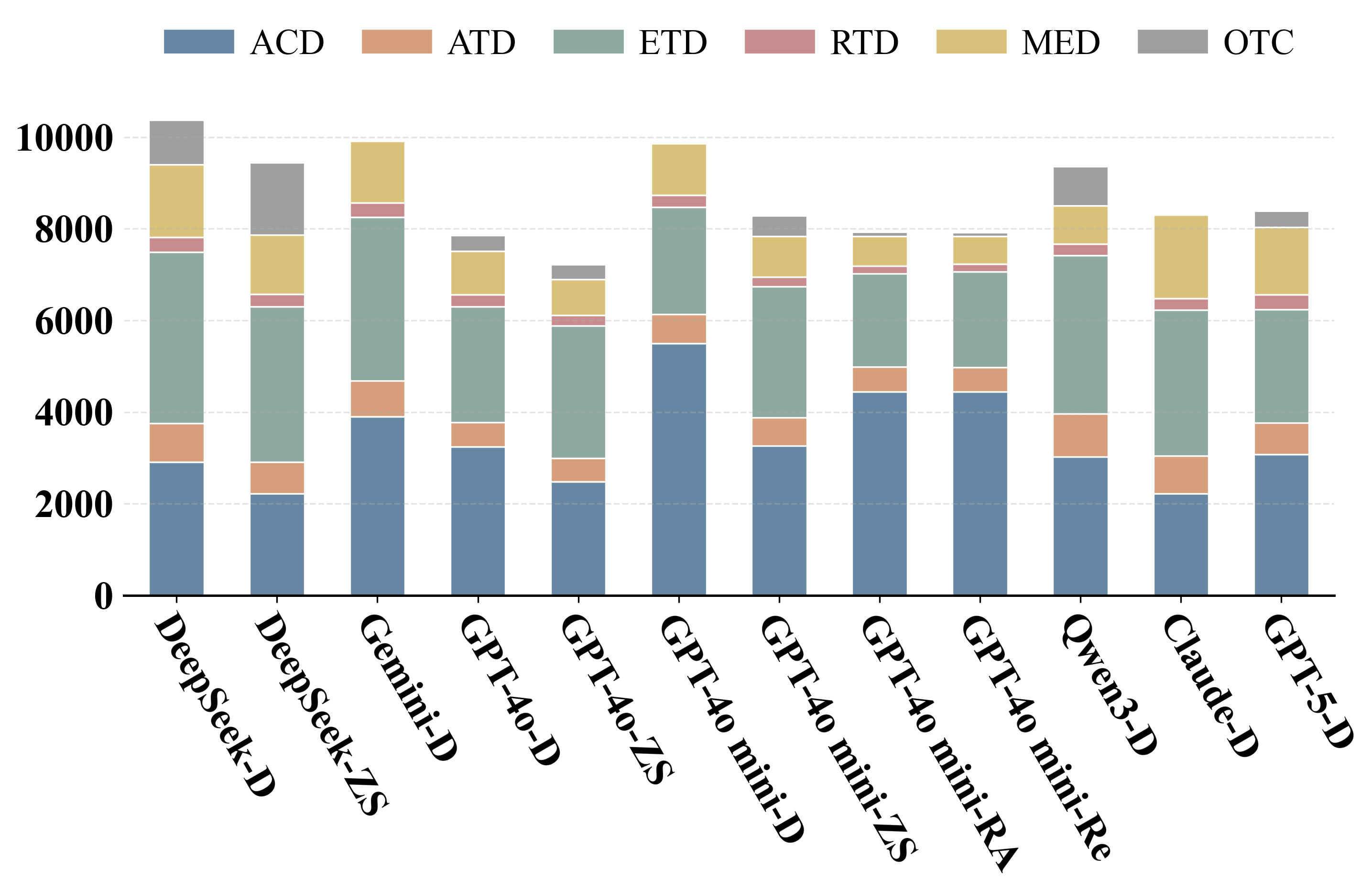}
        \caption{Easy Mode}
        \label{fig:easy_Cost}
    \end{subfigure}
    
    \vspace{1em} 

    \begin{subfigure}[b]{0.48\textwidth}
        \centering
        \includegraphics[width=\textwidth]{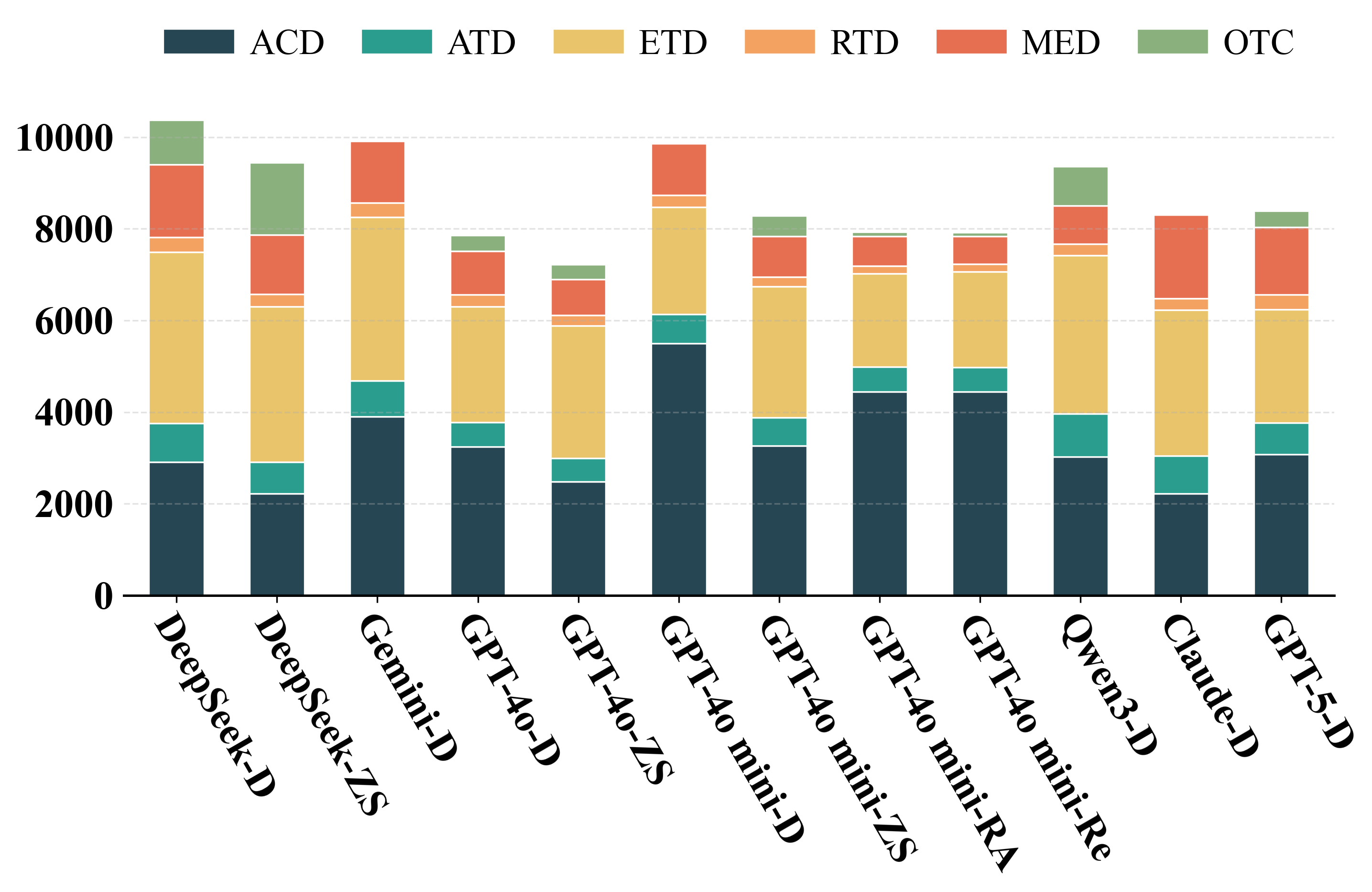}
        \caption{Medium Mode}
        \label{fig:medium_Cost}
    \end{subfigure}
    \hfill
    \begin{subfigure}[b]{0.48\textwidth}
        \centering
        \includegraphics[width=\textwidth]{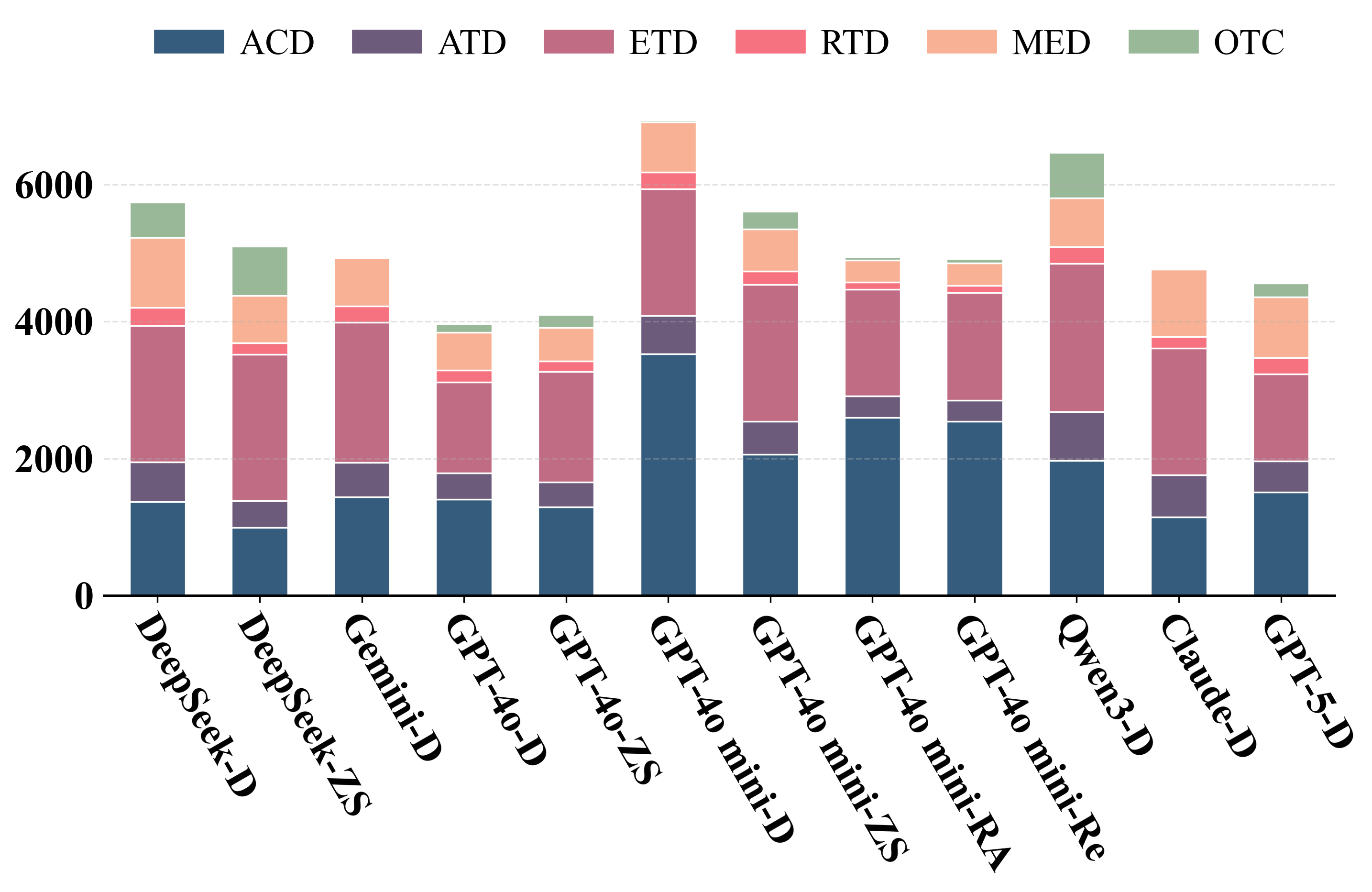}
        \caption{Hard Mode}
        \label{fig:hard_Cost}
    \end{subfigure}
    
    \caption{Comparison of Cost Composition across different difficulty levels. (a) Overall dataset, (b) Easy Mode, (c) Medium Mode, and (d) Hard Mode.}
    \label{fig:full_comparison_Cost}
\end{figure*}

\noindent\textbf{Model backbones.}
\begin{itemize}

\item \textbf{GPT-4o (OpenAI):} A general-purpose proprietary model that serves as a strong reference for instruction following and structured itinerary generation, and is used to characterize high-capability planning performance under TravelEval constraints.  
\item \textbf{GPT-4o-mini (OpenAI):} A lightweight proprietary model representing a cost-efficient deployment tier; we include it to quantify how four mainstream prompting strategies (Direct, Zero-Shot CoT, ReAct, Reflexion) transfer to a smaller model and how strategy-induced gains behave under limited capacity.  
\item \textbf{GPT-5-chat (OpenAI):} A newer-generation proprietary model included to test whether improved reasoning and planning competence translates into better constraint satisfaction and feasibility when generating multi-day itineraries.  
\item \textbf{Claude-4.5 Sonnet (Anthropic):} A leading proprietary model often strong at long-form organization and instruction adherence; it provides a complementary backbone to examine whether different alignment and generation styles affect itinerary consistency and preference compliance.  
\item \textbf{Gemini-2.0-Flash (Google):} A fast proprietary model selected to represent a latency-oriented regime; it helps assess whether responsive models can still produce coherent, feasible schedules under the same structured output requirements.  
\item \textbf{DeepSeek-Chat (DeepSeek-AI):} A competitive open-source model included to broaden backbone diversity and reproducibility; its strong step-wise reasoning behavior is useful for probing how explicit reasoning interacts with travel-specific constraints such as temporal ordering and budget allocation.  
\item \textbf{Qwen3-8B (Alibaba):} A compact open-source instruction model used to represent smaller open models; it helps examine robustness of itinerary structure and constraint handling under tighter capacity while keeping evaluation accessible.
\end{itemize}

\begin{table*}[htbp]
  \centering
  \caption{Detailed Information of the Evaluated LLMs in this Paper.}
  \label{tab:model_details}
  \begin{tabular}{cccc} 
        \toprule
        \textbf{Model} & \textbf{Open-source} & \textbf{Version} & \textbf{Institution} \\
        \midrule
        \textbf{GPT-4o} & No & 20241120 & OpenAI \\
        \rowcolor{rowblue}
        \textbf{GPT-4o-mini} & No & 20240718 & OpenAI \\
        \textbf{GPT-5-chat} & No & 20250807 & OpenAI \\
        \rowcolor{rowblue}
        \textbf{Claude 4.5 Sonnet} & No & 20250929 & Anthropic \\
        \textbf{Gemini 2.0 Flash} & No & 20250205 & Google \\
        \rowcolor{rowblue}
        \textbf{DeepSeek-Chat} & Yes & DeepSeek-chat/DeepSeekV3.1 & DeepSeek-AI  \\
        \textbf{Qwen3-8B} & Yes & Qwen/Qwen3-8B & Alibaba  \\
        \bottomrule
  \end{tabular}
\end{table*}

\noindent\textbf{Reasoning strategies.}
\begin{itemize}
  \item \textbf{Direct (D):} A single-pass prompting setup that directly generates the final travel plan without an explicit intermediate reasoning trace.
  \item \textbf{Zero-Shot CoT (ZS):} A prompting setup that requests an explicit step-by-step reasoning process before emitting the final structured itinerary, intended to improve global coherence and constraint coverage in multi-day planning.
  \item \textbf{ReAct (RA):} An iterative prompting setup following a \textit{Thought--Action--Observation} pattern for a fixed number of rounds, where intermediate self-checks and refinements are used to adjust attraction ordering, resolve conflicts among constraints, and reduce local inconsistencies.
  \item \textbf{Reflexion (Re):} A self-reflection based revision setup that critiques an initial draft and then regenerates an improved plan, aiming to reduce repeated errors and strengthen feasibility and preference alignment in the final structured output.
\end{itemize}
The prompts are shown in Appendix~\ref{sec:Appendix_C}.


\subsection{Baseline Construction}
\label{sec:Appendix_A4}

Baseline is constructed as the evaluation reference method.

\textbf{Design rationale:} from the perspective of metrics design, the baseline should strictly satisfy budget constraints defined in the benchmark while achieving near-optimal performance across the evaluation metrics whenever possible.

\begin{itemize}
    \item Extract budget, number of travelers (including people's composition), and explicit user preference information from the user query. Based on the extracted preferences, filter attractions that match the user's preferences and rank them in descending order by attraction score, selecting the top $X$ attractions as candidate POIs. In parallel, construct an additional candidate set by ranking attractions purely by score without considering preference information, and similarly select the top $X$ attractions. The total number of candidate attractions is thus $2X$, where $2X$ is determined by the attraction density associated with the traveler composition multiplied by the total number of travel days.

    \item Perform spatial clustering on the candidate attractions. Attractions within the same cluster are scheduled on the same day. Except for the first and last days (whose clusters each contain only one attraction), the remaining clusters are assigned an average number of attractions. Based on the coordinates of the attraction clusters on the first and last days, select the nearest transportation hubs as the destination and departure points for intercity transportation, respectively.

    \item Select the hotel that minimizes the aggregated distance to all selected Points of Interest (POIs) as the candidate accommodation. If accommodation preferences are specified, select the hotel that both satisfies the preferences and has the minimum aggregated distance to the selected POIs.

    \item Design the daily visiting order and specific visiting times for attractions by jointly considering attraction opening hours and recommended visiting durations. The visiting sequence is required to satisfy local path optimality, i.e., minimizing the total intra-day travel distance.

    \item Select dining locations along the optimized travel paths by choosing restaurant POIs that are closest to the planned routes while satisfying user dietary preferences.

    \item Supplement the itinerary with intercity and intracity transportation options that satisfy user preferences and remain within the specified budget constraints.

    \item Output the complete travel plan in the predefined JSON schema.
\end{itemize}

\textbf{Note:} Approach A permits moderate budget overruns for travel plans, prioritizing the maximization of users' attraction experience and full alignment with their preferences. For Approach B, all components of the baseline design must strictly adhere to the budget constraint. If a plan exceeds the budget, partial user preferences need to be gradually abandoned until the budget is balanced. Otherwise, clear feedback should be given when it is truly impossible to meet the requirements.

\subsection{Experimental Environment.} 
\label{sec:Appendix_A4}
All experiments were conducted on a standard cloud server with Ubuntu 22.04 and Python $\geq$3.11. All models were accessed via their official APIs. Each model-strategy combination was evaluated on our benchmark, and the resulting plans were automatically scored by our multi-dimensional evaluation framework.

\section{Supplementary Experimental Results and Case Studies}

This appendix provides (i) the full metric results across all model--strategy configurations,
(ii) the metric abbreviation map, and (iii) additional case studies and diagnostic visualizations
that complement the main-text analyses.

\begin{table*}[t]
\centering
\scriptsize
\setlength{\tabcolsep}{2.2pt}
\renewcommand{\arraystretch}{1.2}
\caption{Full results across model/strategy configurations. \textbf{Dir} and \textbf{ZS-CoT} denote Direct and Zero-Shot Chain-of-Thought prompting, respectively. Arrows indicate the direction of better performance: $\uparrow$ higher is better, $\downarrow$ lower is better.}
\label{tab:full_results_overview}
    \renewcommand{\arraystretch}{1.5} 
    
    \resizebox{\textwidth}{!}{%
        \setlength{\tabcolsep}{3.5pt} 
        
        \begin{tabular}{l cccccccccccccc}
            \toprule
            \multirow{2.5}{*}{\textbf{Metric}} & 
            \multicolumn{2}{c}{\textbf{Deepseek}} & 
            \textbf{Gemini 2.0} & 
            \multicolumn{2}{c}{\textbf{GPT-4o}} & 
            \multicolumn{4}{c}{\textbf{GPT-4o mini}} & 
            \textbf{Qwen3-8b} & 
            \textbf{Claude 4.5} & 
            \textbf{GPT-5} & 
            \multirow{2.5}{*}{\shortstack{\textbf{Approach A}}} &
            \multirow{2.5}{*}{\shortstack{\textbf{Approach B}}} \\ 
            
            \cmidrule(lr){2-3} \cmidrule(lr){4-4} \cmidrule(lr){5-6} \cmidrule(lr){7-10} \cmidrule(lr){11-11} \cmidrule(lr){12-12} \cmidrule(lr){13-13}
            
             & Dir & ZS-CoT & Dir & Dir & ZS-CoT & Dir & ZS-CoT & ReAct & Reflexion & Dir & Dir & Dir & & \\ 
            \midrule

            \rowcolor{graybg} \multicolumn{15}{c}{\textbf{Accuracy Dimension}} \\
            CCD $\downarrow$   & 0.1257 & 0.2082 & 0.1363 & 0.1385 & 0.1848 & 0.2789 & 0.3288 & 0.2364 & 0.2308 & 0.1911 & \textbf{0.0691} & 0.0995 & 0.0005 & 0.0013 \\
            FAR $\downarrow$   & 0.0437 & 0.0426 & 0.0364 & 0.0367 & 0.0537 & 0.039 & 0.0528 & 0.0879 & 0.0951 & 0.0249 & \textbf{0.0292} & 0.0451 & 0 & 0 \\
            VROH $\downarrow$  & 0.0454 & 0.0360 & 0.0443 & 0.0327 & 0.0461 & 0.0446 & 0.0626 & 0.0431 & 0.0438 & 0.0507 & 
            \textbf{0.0316} & 0.0574 & 0.2504 & 0.0046 \\
            FRTC $\downarrow$  & 0 & 0 & 0 & 0 & 0 & 0 & 0 & 0 & 0 & 0 &0 & 0 & 0 & 0\\
            ITD $\downarrow$   & 0.1450 & 0.1572 & 0.1718 & 0.1866 & 0.2008 & 0.3233 & 0.3389 & 0.3258 & 0.3288 & 0.3281 &
            \textbf{0.1401} & 0.1621 & 0.4488 & 0.8605 \\
            PD $\downarrow$    & 0.6214 & 0.6103 & 0.6441 & 0.7420 & 0.6558 & 0.6546 & 0.4158 & 0.7300 & 0.7349 & 0.5974 & \textbf{0.3039} & 0.7135 & 0 & 0\\
            
            \rowcolor{graybg} \multicolumn{15}{c}{\textbf{Compliance Dimension}} \\
            BCS $\uparrow$     & 0.1794 & 0.2875 & 0.2334 & \textbf{0.3826} & 0.3758 & 0.1701 & 0.3199 & 0.3694 & 0.3701 & 0.1680 & 0.3110& 0.2412 & 0.6273 & 0.9217\\
            TCS $\uparrow$     & \textbf{1} & 0.9878 & \textbf{1} & 0.9948 & 0.9946 & \textbf{1} & 0.9982 & 0.9945 & 0.9935 & \textbf{1} &0.9985 & 0.9921 & 1 & 1\\
            HCS $\uparrow$     & 0.6464 & 0.6124 & 0.4659 & 0.4835 & 0.4728 & 0.5872 & 0.5951 & 0.6077 & 0.6037 & 0.3718 & \textbf{0.6860} & 0.5689 & 1 & 1\\
            PAS-A $\uparrow$   & 0.5277 & 0.4852 & 0.5269 & 0.5113 & 0.4292 & 0.1169 & 0.1718 & 0.1604 & 0.1589 & 0.4870 & \textbf{0.5699}& 0.5018 & 0.6681 & 0.6748\\
            PAS-T $\uparrow$   & 0.9983 & 0.9643 & 0.9919 & \textbf{1} & \textbf{1} & 0.9929 & 0.9965 & 0.9890 & 0.9888 & \textbf{1} & 0.9955 & 0.9956 & 0.7524 & 0.7826\\
            PAS-C $\uparrow$   &\textbf{0.8354} & 0.7552 & 0.7756 & 0.6730 & 0.6253 & 0.6067 & 0.5960 & 0.5674 & 0.5579 & 0.5996 & 0.7991& 0.6935 & 0.9036 & 0.5652\\
            TTCS $\uparrow$    & 0.0147 & 0.0235 & 0.009 & 0.0565 & 0.0545 & 0.0461 & 0.0561 & 0.2264 & \textbf{0.2421} & 0.0572 & 0.0089 & 0.0159 & 0.0096 & 0.4487\\

            \rowcolor{graybg} \multicolumn{15}{c}{\textbf{Temporality Dimension}} \\
            STR $\uparrow$     & 0.7023 & 0.7338 & 0.7476 & 0.6868 & 0.6956 & 0.7294 & \textbf{0.7602} & 0.6783 & 0.6631 & 0.7313 & 0.6820 & 0.6959 & 0.4979 & 0.7677\\
            DTU $\uparrow$     & 0.2551 & 0.2683 & \textbf{0.2715} & 0.2176 & 0.2076 & 0.2363 & 0.2528 & 0.2218 & 0.2132 & 0.2346 & 0.2597 & 0.2217 & 0.1212 & 0.3029\\
            OTU $\uparrow$     & 0.0528 & 0.0536 & 0.0541 & 0.0388 & 0.0357 & 0.0404 & 0.0379 & 0.0289 & 0.0276 & 0.0314 & \textbf{0.0573} & 0.0441 & 0.0608 & 0.0649\\

            \rowcolor{graybg} \multicolumn{15}{c}{\textbf{Spatiality Dimension}} \\
            SSR $\downarrow$     & 0.1257 & 0.0930 & 0.1298 & 0.0616 & 0.0390 & 0.0353 & 0.0266 & \textbf{0.0109} & 0.0112 & 0.0295 &0.1058& 0.0907 & 0.0970 & 0.0660\\
            CSM-P90 $\downarrow$ & 3.0156 & 3.095 & 3.0189 & 3.7863 & 3.2558 & 3.5632 & 3.2657 & 1.6321 & \textbf{1.5101} & 2.9278 & 2.9327 & 3.3794 & 1.8498 & 1.5222\\
            CSM-P95 $\downarrow$ & 4.0396 & 4.5713 & 3.9585 & 4.8575 & 4.5672 & 4.9057 & 4.3486 & 2.1094 & \textbf{1.8967} & 3.8200 & 4.3765& 4.7063 & 5.6064 & 2.4045\\

            \rowcolor{graybg} \multicolumn{15}{c}{\textbf{Economy Dimension}} \\
            BE $\uparrow$      & 8.8035 & 9.1860 & 9.0362 & 7.5459 & 6.9124 & 4.7187 & 5.6968 & 4.3266 & 4.2641 & 4.0518 & \textbf{10.752} & 7.9269 & 16.274 & 22.192\\
            ACD    & 2045.7 & \textbf{1523.0} & 2643.3 & 2273.1 & 1825.1 & 4620.4 & 2639.3 & 3569.9 & 3560.1 & 2559.5 & 1589.7& 2244.8 &  1294.2 & 474.66\\
            ATD    & 719.23 & 547.16 & 667.20 & 472.57 & 473.84 & 592.14 & 558.56 & 450.62 & \textbf{433.59} & 831.48 & 731.23& 585.21 & 142.52 & 316.27\\
            ETD    & 2867.5 & 2778.4 & 2896.9 & 2009.3 & 2353.2 & 2035.4 & 2345.7 & \textbf{1732.0} & 1744.6 & 2932.2 & 2653.5& 1894.3 &  2668.4 & 1736.47 \\
            RTD   & 278.80 & 206.40 & 263.27 & 209.98 & 186.49 & 238.45 & 190.53 & 132.59 & \textbf{131.25} & 248.22 &195.47 & 266.33 & 206.80 & 203.86\\
            MED    & 1268.6 & 908.89 & 998.50 & 735.96 & 612.33 & 893.50 & 666.24 & 442.81 & \textbf{426.51} & 755.18 & 1341.7& 1153.0 &  1426.5 & 137.29 \\
            OTD     & 769.81 & 1030.9 & \textbf{10.033} & 249.02 & 287.96 & 31.129 & 374.01 & 73.740 & 70.033 & 747.51 & 25.542 & 304.92 & 0
            & 0  \\

            \rowcolor{graybg} \multicolumn{15}{c}{\textbf{Utility Dimension}} \\
            EDI $\uparrow$     & 0.6152 & 0.6161 & 0.6176 & 0.5627 & 0.5443 & 0.5627 & 0.5465 & 0.4505 & 0.4364 & 0.5271 & \textbf{0.6235} & 0.6112 & 0.6547 & 0.5829\\
            ADS $\uparrow$     & 0.6275 & 0.6017 & 0.5989 & 0.4692 & 0.4272 & 0.4326 & 0.3968 & 0.3296 & 0.3124 & 0.3500 & \textbf{0.6720} & 0.5333 & 0.7649 & 0.5129\\
            AQE $\uparrow$     & 0.9913 & 0.9788 & 1.0061 & 0.7251 & 0.7028 & 0.713 & 0.7435 & 0.5618 & 0.5411 & 0.6781 & \textbf{1.0298}& 0.8448 & 0.9882 & 0.7479\\
            Profit $\uparrow$  & 5.7182 & 5.5473 & 5.7102 & 5.9691 & 5.9576 & \textbf{5.9661} & 5.7625 & 5.5359 & 5.4862 & 5.7353 &5.4352& 5.6541 & 5.9979 & 5.5015\\
            \bottomrule

\end{tabular}%
}
\end{table*}


\begin{table*}[htbp] 
    \centering
    \small
    \caption{Full Results of Supplementary Experiments (Reformatted)}
    \label{tab:Full_Results_of_Supplementary_Experiments}
    \renewcommand{\arraystretch}{1.2} 
    
    \begin{tabular*}{\textwidth}{@{\extracolsep{\fill}} l ccccccccc @{}} 
        \toprule 
        
        \multirow{2}{*}{\textbf{Metric}} & 
        \multicolumn{2}{c}{\textbf{DeepSeek}} & 
        \multicolumn{3}{c}{\textbf{GPT-5}} & 
        \textbf{Claude} & 
        \multicolumn{3}{c}{\textbf{Qwen}} \\
        
        \cmidrule(lr){2-3} \cmidrule(lr){4-6} \cmidrule(lr){7-7} \cmidrule(lr){8-10}
        
        & RA & Re & ZS & RA & Re & ZS & ZS & RA & Re \\
        \midrule
        
        \rowcolor{gray!15} \multicolumn{10}{c}{\textbf{Accuracy Dimension}} \\
        CCD $\downarrow$ & 0.1036 & 0.1017 & 0.1811 & 0.1235 & 0.1265 & 0.1030 & 0.2156 & 0.1577 & 0.1683 \\ 
        FAR $\downarrow$ & 0.0594 & 0.0570 & 0.0456 & 0.0392 & 0.0450 & 0.0443 & 0.0335 & 0.0236 & 0.0234 \\ 
        VROH $\downarrow$ & 0.0150 & 0.0152 & 0.0541 & 0.0534 & 0.0509 & 0.0357 & 0.0572 & 0.0641 & 0.0644 \\ 
        FRTC $\downarrow$ & 0 & 0 & 0 & 0 & 0 & 0 & 0 & 0 & 0 \\ 
        ITD $\downarrow$ & 0.1569 & 0.1582 & 0.1499 & 0.1495 & 0.1496 & 0.1492 & 0.3409 & 0.2778 & 0.2801 \\ 
        PD $\downarrow$ & 0.5872 & 0.5859 & 0.6232 & 0.8063 & 0.8058 & 0.4113 & 0.5066 & 0.8060 & 0.6813 \\ 
        
        \rowcolor{gray!15} \multicolumn{10}{c}{\textbf{Compliance Dimension}} \\
        BCS $\uparrow$ & 0.4719 & 0.4727 & 0.4125 & 0.3648 & 0.3703 & 0.4613 & 0.3099 & 0.3915 & 0.2909 \\ 
        TCS $\uparrow$ & 0.9705 & 0.9684 & 0.9936 & 0.9972 & 0.9979 & 0.9952 & 0.9942 & 0.9987 & 0.9987 \\ 
        HCS $\uparrow$ & 0.5815 & 0.5805 & 0.5829 & 0.5127 & 0.5165 & 0.5530 & 0.4990 & 0.4251 & 0.4208 \\ 
        PAS-A $\uparrow$ & 0.4537 & 0.4540 & 0.4383 & 0.3685 & 0.3714 & 0.5030 & 0.2076 & 0.2842 & 0.2857 \\ 
        PAS-T $\uparrow$ & 0.8806 & 0.8750 & 0.9807 & 0.9830 & 0.9819 & 0.9869 & 0.9961 & 0.9884 & 0.9870 \\ 
        PAS-C $\uparrow$ & 0.6994 & 0.6997 & 0.6648 & 0.6711 & 0.6649 & 0.7187 & 0.5838 & 0.5827 & 0.5831 \\ 
        TTCS $\uparrow$ & 0.0253 & 0.0230 & 0.0249 & 0.0292 & 0.0406 & 0.0072 & 0.0400 & 0.0866 & 0.0870 \\ 
        
        \rowcolor{gray!15} \multicolumn{10}{c}{\textbf{Temporality Dimension}} \\
        STR $\uparrow$ & 0.6492 & 0.6493 & 0.6947 & 0.7215 & 0.7136 & 0.7043 & 0.7407 & 0.7396 & 0.7399 \\ 
        DTU $\uparrow$ & 0.2267 & 0.2262 & 0.2278 & 0.2331 & 0.2303 & 0.2546 & 0.2397 & 0.2516 & 0.2536 \\ 
        OUT $\uparrow$ & 0.0504 & 0.0503 & 0.0409 & 0.0411 & 0.0406 & 0.0547 & 0.0371 & 0.0332 & 0.0332 \\ 
        
        \rowcolor{gray!15} \multicolumn{10}{c}{\textbf{Spatiality Dimension}} \\
        SSR $\downarrow$ & 0.1082 & 0.1089 & 0.0567 & 0.0737 & 0.0680 & 0.0905 & 0.0269 & 0.0205 & 0.0205 \\ 
        CSM-P90 $\downarrow$ & 3.0023 & 2.9912 & 3.2935 & 3.1679 & 3.1696 & 3.3060 & 2.9980 & 2.1499 & 2.1590 \\ 
        CSM-P95 $\downarrow$ & 3.8507 & 3.8501 & 4.8097 & 4.3760 & 4.3798 & 4.5684 & 4.1104 & 2.8355 & 2.8480 \\ 
        
        \rowcolor{gray!15} \multicolumn{10}{c}{\textbf{Economy Dimension}} \\
        BE $\uparrow$ & 9.9504 & 9.8806 & 8.0591 & 7.2305 & 7.1041 & 10.958 & 6.3698 & 5.3970 & 4.6926 \\ 
        ACD & 1831.9 & 1831 & 1987 & 2384 & 2366 & 1272 & 1870 & 1974 & 1974 \\ 
        ATD & 448.16 & 454.93 & 518.82 & 514.15 & 505.40 & 469.29 & 791.19 & 691.91 & 691.50 \\ 
        ETD & 2948.2 & 2959.0 & 2525.6 & 2257.9 & 2239.4 & 2502.3 & 2851.6 & 2249.3 & 3305.2 \\ 
        RTD & 192.81 & 194.09 & 208.00 & 219.1 & 214.08 & 176.17 & 291.45 & 245.90 & 246.61 \\ 
        MED & 1059.3 & 1072.3 & 569.09 & 657.36 & 636.27 & 972.59 & 730.44 & 606.38 & 607.96 \\ 
        OTD & 0 & 0 & 0 & 0 & 0 & 0 & 0 & 0 & 0 \\ 
        
        \rowcolor{gray!15} \multicolumn{10}{c}{\textbf{Utility Dimension}} \\
        EDI $\uparrow$ & 0.6135 & 0.6162 & 0.5042 & 0.4857 & 0.4830 & 0.6262 & 0.3936 & 0.3351 & 0.3350 \\ 
        ADS $\uparrow$ & 0.9394 & 0.9401 & 0.8218 & 0.8204 & 0.8101 & 1.0006 & 0.7705 & 0.7142 & 0.7121 \\ 
        AQE $\uparrow$ & 0.6126 & 0.6151 & 0.5926 & 0.5845 & 0.5755 & 0.6028 & 0.5530 & 0.5083 & 0.5080 \\ 
        Profit $\uparrow$ & 5.5156 & 5.5190 & 5.6358 & 5.6827 & 5.6486 & 5.4902 & 5.7186 & 5.5659 & 5.5565 \\ 
        
        \bottomrule 
    \end{tabular*}
\end{table*}

\subsection{Full Benchmark Results}

\noindent Table~\ref{tab:full_results_overview} reports the complete results of all evaluated model-strategy combinations, together with our two rule-based baselines, across six \texttt{TravelEval} dimensions and their fine-grained metrics. Overall, no single model or prompting strategy consistently performs well across all dimensions, and the baselines serve as reference points for feasibility- and constraint-oriented performance.

To guarantee fair and controlled comparisons under computational resource constraints, we first benchmark all candidate models under the Direct prompting strategy. We further conducted exhaustive experiments covering all four prompting strategies on GPT-4o mini to fully verify their practical effects. This experimental framework enables valid, unified comparisons among different prompting paradigms. Our primary results reveal that advanced sophisticated prompting strategies only bring marginal performance improvements for travel planning tasks. To solidify this finding, we supplemented a series of key auxiliary experiments (summarized in Table~\ref{tab:Full_Results_of_Supplementary_Experiments}). The outputs consistently corroborate our core observation that complex prompting schemes deliver limited overall gains on travel planning.

We also carried out a small-scale preliminary validation of tool-augmented agents based on locally deployed Claude Code. Over 250 randomly sampled queries were tested in this experiment, and the outcomes are listed in Table~\ref{tab:claude_code_full}. The results reveal steady gains across multiple core metrics including BCS, which demonstrates that our benchmark framework can be seamlessly extended to the evaluation of tool-based travel agents.

Figure~\ref{fig:full_comparison_Cost} shows the cost composition across difficulty levels, decomposing total expenditure into accommodation (ACD), attractions (ATD), intercity transportation (ETD), intracity transportation (RTD), meals (MED), and other costs (OTC). Across all settings, intercity transportation and accommodation account for the largest shares of total cost, indicating that long-distance travel choices and accomodation decisions dominate budget allocation regardless of model or strategy. As difficulty increases from Easy to Hard, total spending decreases across all components, but the overall cost distribution pattern remains largely consistent. Variations across models and prompting strategies are mainly in accommodation and other costs, whereas transportation-related components exhibit less fluctuation. 

\begin{figure*}
    \centering
    \includegraphics[width=\textwidth]{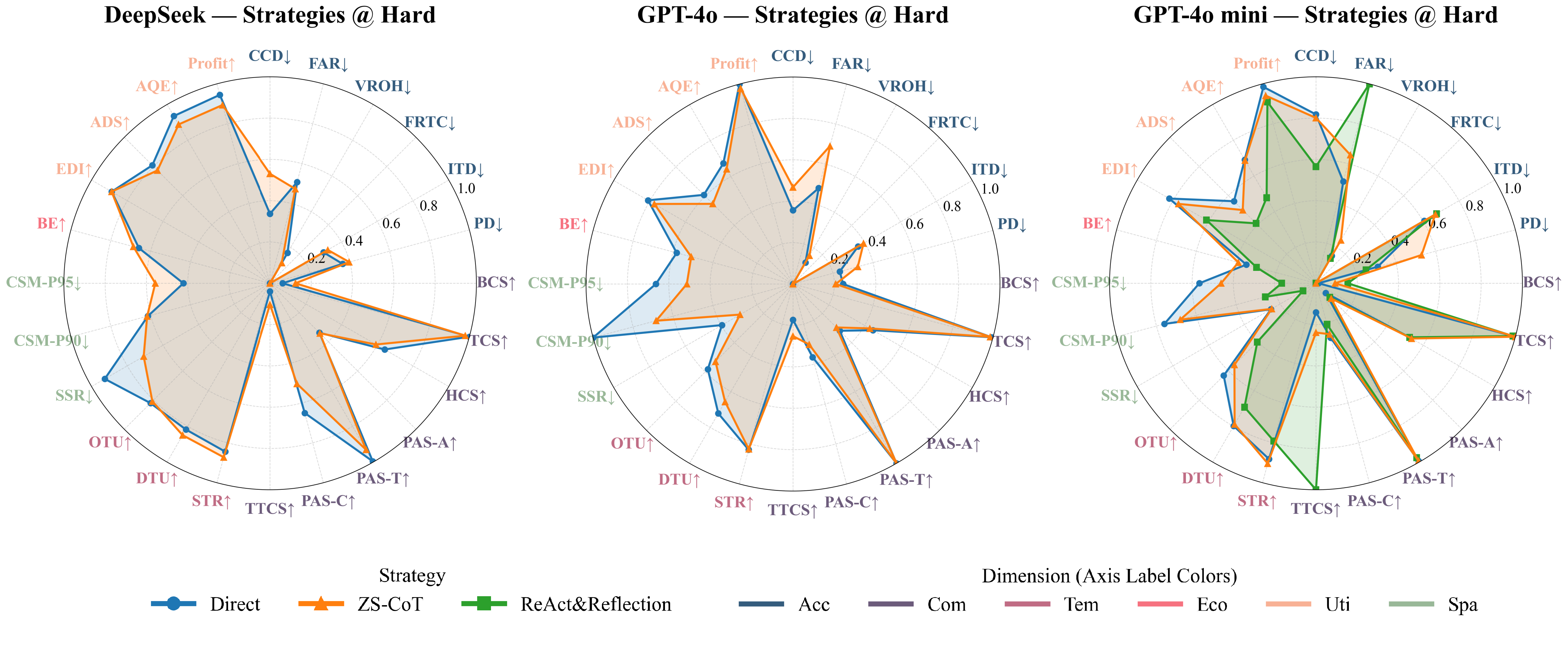}
    \caption{Comparative analysis of Reasoning Strategies on hard queries. Performance is normalized across 26 metrics, grouped and color-coded by evaluation dimension. }
    \label{fig:strategy_comparison_hard}
\end{figure*}

In addition, we visualize the strategy-wise performance of Deep\-Seek, GPT-4o, and GPT-4o-mini under the Hard setting using a radar chart (Figure~\ref{fig:strategy_comparison_hard}) as a supplement to Section~\ref{sec:MainResults}. Consistent with the main results, increasing strategy complexity does not lead to a consistent overall performance gain: while certain metrics improve, more complex strategies often underperform Direct prompting across multiple evaluation metrics.

\subsection{Case Studies}
\label{app:case_study}

\begin{figure*}
    \centering
    \includegraphics[width=\textwidth]{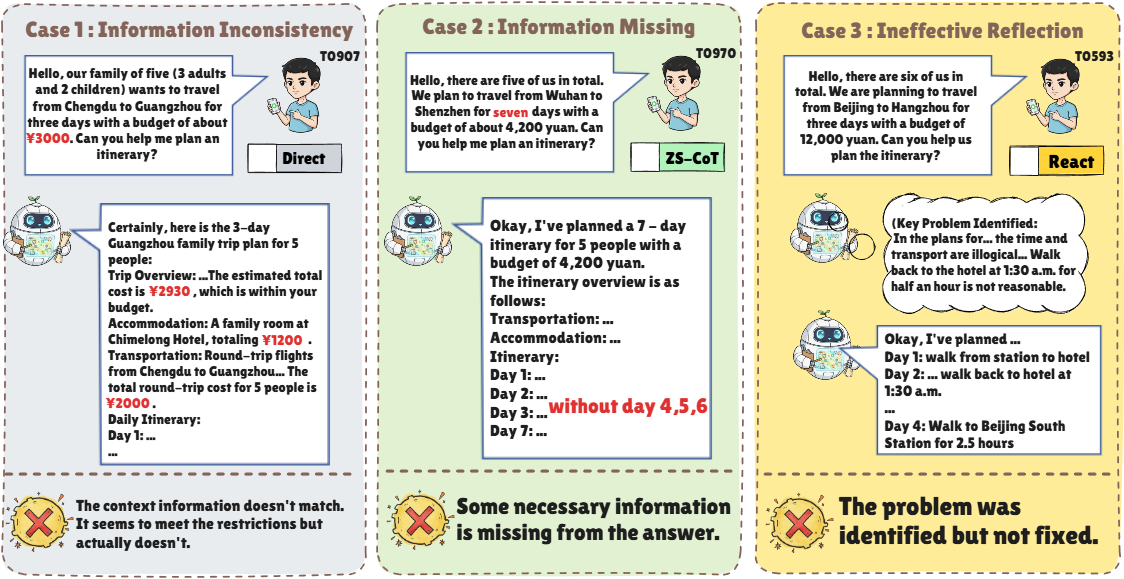}
    \caption{Case Study}
    \label{fig:casestudy}
\end{figure*}

\noindent Figure~\ref{fig:casestudy} visualizes representative failure cases discussed below.

\paragraph{Case 1: Numerical inconsistency and false compliance (\texttt{T0907 (Hard)-Gemini Direct}).}
The model claimed that the Travel Plan satisfied the 3{,}000 RMB budget, reporting a total Cost of 2{,}930 RMB, yet the itemized breakdown summed to 5{,}420 RMB. Intercity flight Cost alone (4{,}000 RMB) already exceeded the entire budget. This case demonstrates a fundamental failure in multi-step numerical calculation and cross-component consistency checking, leading to incorrect budget compliance.

\paragraph{Case 2: Strategy-induced failure (\texttt{T0970 (Hard)-GPT-4o ZS-CoT} vs. \texttt{Direct}).}
Under Zero-Shot CoT, the agent produced a critically incomplete Travel Plan, omitting Days 4--6 in a 7-day request, while still reporting a suspicious exact match between the computed Cost and the budget (4{,}200 RMB) despite the missing days. In contrast, a Direct strategy produced a complete plan within budget (4{,}130 RMB). This case illustrates that adding a reasoning step can be detrimental when the reasoning process itself is flawed, amplifying error accumulation and plan incompleteness.

\begin{table*}[htbp]
    \centering
    \caption{Ablation Study Results}
    \label{tab:ablation_study}
     \setlength{\tabcolsep}{2.5pt}
    \renewcommand{\arraystretch}{1.4} 
    
    \resizebox{\textwidth}{!}{
    \begin{tabular}{@{} l cccccccccccccc @{}}
        \toprule
        
        & \textbf{DeepSeek-} & \textbf{DeepSeek-} & \textbf{Gemini-} & \textbf{GPT-4o-} & \textbf{GPT-4o-} & \textbf{GPT-4o} & \textbf{GPT-4o} & \textbf{GPT-4o} & \textbf{GPT-4o} & \textbf{GPT5-}& \textbf{Claude-} & \textbf{Qwen-} & \textbf{Approach} & \textbf{Approach} \\
        
        \textbf{Model} & \textbf{D} & \textbf{ZS} & \textbf{D} & \textbf{D} & \textbf{ZS} & \textbf{mini-D} & \textbf{mini-ZS} & \textbf{mini-RA} & \textbf{mini-Re} & \textbf{D} & \textbf{D} & \textbf{D} & \textbf{A} & \textbf{B} \\
        \midrule
        
        STR 
        & 0.7023 & 0.7338 & 0.7476 & 0.6868 & 0.6956 & 0.7294 & 0.7602 & 0.6783 & 0.6631 & 0.6959 & 0.6820 & 0.7313 & 0.4979 & 0.7677 \\
        \addlinespace 
        
        STR without Queuing 
        & 0.8683 & 0.8744 & 0.9116 & 0.8497 & 0.8476 & 0.8783 & 0.9001 & 0.7970 & 0.7794 & 0.8485 & 0.8138 & 0.8841 & 0.6616 & 0.9503 \\
        \addlinespace
        
        DTU 
        & 0.2550 & 0.2683 & 0.2715 & 0.2176 & 0.2076 & 0.2363 & 0.2528 & 0.2218 & 0.2132 & 0.2217 & 0.2597 & 0.2346 & 0.1212 & 0.3029 \\
        \addlinespace
        
        DTU without Queuing 
        & 0.3087 & 0.3132 & 0.3248 & 0.2642 & 0.2481 & 0.2811 & 0.2934 & 0.2564 & 0.2464 & 0.2645 & 0.3032 & 0.2784 & 0.1563 & 0.3552 \\
        \addlinespace
        
        OUT 
        & 0.0528 & 0.0536 & 0.0541 & 0.0388 & 0.0357 & 0.0404 & 0.0379 & 0.0289 & 0.0276 & 0.0441 & 0.0573 & 0.0314 & 0.0608 & 0.0649 \\
        \addlinespace
        
        OUT without Queuing 
        & 0.0660 & 0.0646 & 0.0665 & 0.0487 & 0.0441 & 0.0494 & 0.0458 & 0.0352 & 0.0336 & 0.0544 & 0.0689 & 0.0389 & 0.0825 & 0.0777 \\
        \addlinespace
        
        Distances with API
        & 11.2526 & 11.2299 & 10.6174 & 14.6933 & 14.0636 & 17.6460 & 17.6443 & 17.2488 & 17.4917 & 11.1742 & 9.2028 & 13.1541 & 13.3085 & 23.8097 \\
        \addlinespace
        
        Distances with Geodesic
        & 7.6632 & 7.7087 & 7.3298 & 10.4366 & 9.9340 & 12.6001 & 12.7404 & 12.3841 & 12.5806 & 7.6489 & 6.1543 & 9.1459 & 9.0460 & 17.0517 \\
        
        \bottomrule
    \end{tabular}
    }
\end{table*}

\begin{table}[t] 
    \centering
    \small 
    \caption{The results of Claude Code's experiments}
    \label{tab:claude_code_full}
    \renewcommand{\arraystretch}{1.15} 
    
    \begin{tabular*}{\linewidth}{@{\extracolsep{\fill}} llcccc @{}} 
        \toprule 
        \textbf{Dimension} & \textbf{Metric} & \textbf{Overall} & \textbf{Easy} & \textbf{Medium} & \textbf{Hard} \\
        \midrule 
        
        \multirow{6}{*}{\textbf{Accuracy}} 
        & CCD $\downarrow$ & 0.0407 & 0.0380 & 0.0449 & 0.0383 \\ 
        & FAR $\downarrow$ & 0 & 0 & 0 & 0 \\ 
        & VROH $\downarrow$ & 0.0235 & 0.0275 & 0.0273 & 0.0173 \\ 
        & FRTC $\downarrow$ & 0 & 0 & 0 & 0 \\ 
        & ITD $\downarrow$ & 0.0460 & 0.0419 & 0.0420 & 0.0527 \\ 
        & PD $\downarrow$ & 0.2148 & 0.2333 & 0.2105 & 0.2075 \\ \midrule
        
        \multirow{7}{*}{\textbf{Compliance}} 
        & BCS $\uparrow$ & 0.5703 & 0.6508 & 0.8100 & 0.2800 \\ 
        & TCS $\uparrow$ & 1 & 1 & 1 & 1 \\ 
        & HCS $\uparrow$ & 0.5361 & 0.4603 & 0.5300 & 0.5900 \\ 
        & PAS-A $\uparrow$ & 0.4943 & 1 & 0.3500 & 0.3200 \\ 
        & PAS-T $\uparrow$ & 0.8479 & 1 & 0.8200 & 0.7800 \\ 
        & PAS-C $\uparrow$ & 0.5399 & 1 & 0.6100 & 0.1800 \\ 
        & TTCS $\uparrow$ & 0.0342 & 0.0317 & 0.0300 & 0.0400 \\ \midrule
        
        \multirow{3}{*}{\textbf{Temporality}} 
        & STR $\uparrow$ & 0.6958 & 0.6900 & 0.7028 & 0.6925 \\ 
        & DTU $\uparrow$ & 0.1955 & 0.1754 & 0.1973 & 0.2064 \\ 
        & OTU $\uparrow$ & 0.0503 & 0.0492 & 0.0505 & 0.0509 \\ \midrule
        
        \multirow{3}{*}{\textbf{Spatialilty}} 
        & SSR $\downarrow$ & 0.2830 & 0.2701 & 0.2871 & 0.2871 \\ 
        & CSM-P90 $\downarrow$ & 2.9715 & 2.4953 & 3.0774 & 3.1676 \\ 
        & CSM-P95 $\downarrow$ & 3.4778 & 2.9072 & 3.6475 & 3.6696 \\ \midrule
        
        \multirow{7}{*}{\textbf{Economy}} 
        & BE $\uparrow$ & 13.212 & 11.623 & 12.505 & 14.921 \\ 
        & ACD & 159.15 & 139.32 & 154.39 & 176.40 \\ 
        & ATD & 227.05 & 225.68 & 213.99 & 240.97 \\ 
        & ETD & 3593.1 & 3735.9 & 3427.2 & 3669.1 \\ 
        & RTD & 330.11 & 307.75 & 341.24 & 333.07 \\ 
        & MTD & 431.95 & 432.10 & 417.75 & 446.05 \\ 
        & OTD & 0.0000 & 0.0000 & 0.0000 & 0.0000 \\ \midrule
        
        \multirow{4}{*}{\textbf{Utility}} 
        & EDI $\uparrow$ & 0.5966 & 0.6032 & 0.5840 & 0.6051 \\ 
        & ADS $\uparrow$ & 0.5375 & 0.4976 & 0.5149 & 0.5853 \\ 
        & AQE $\uparrow$ & 0.8033 & 0.7809 & 0.7999 & 0.8209 \\ 
        & Profit $\uparrow$ & 5.6632 & 4.9440 & 5.3539 & 6.4255 \\ 
        \bottomrule 
    \end{tabular*}
\end{table}

\paragraph{Case 3: Ineffective Reflexion and failed refinement (\texttt{T0593 (Medium) -- ReAct/} \texttt{Reflexion}).}
For a G2-level User Query, the Reflexion process correctly identified a critical temporal flaw but failed to correct it in the final structured output: the Travel Plan still contained nonsensical \texttt{ending\_point} times of 01{:}00--01{:}30 (AM) for both days. This case highlights a gap between error identification and effective plan correction, where iterative Reasoning Strategies degrade into ineffective local adjustments under preference-driven complexity.

These cases provide concrete evidence for the limitations discussed: fundamental numerical and consistency issues, potential negative impacts of Reasoning Strategies, and the critical failure to handle complex constraints and trade-offs through effective replanning or optimization.

\subsection{Ablation Study}
\label{app:Alation study}

\noindent Table~\ref{tab:ablation_study} results show that removing the queue modeling module inflates the Scene-Time Ratio (STR), Daily Time Utilization (DTU), and Overall Time Utilization (OUT) by 21.6\%, 19.0\%, and 23.4\% on average, respectively. This finding demonstrates the critical role of queue modeling, as its absence leads to a significant overestimation of real-time utilization efficiency. Furthermore, replacing real road network distances with naive geodesic distances reduces the average inter-attraction distance by 30.1\%, confirming that real route APIs significantly improve the accuracy of spatial rationality evaluation.

\section{JSON Schema and Prompts}
\label{sec:Appendix_C}
All examples and prompts of user query and strategies are as follows.
\twocolumn









\onecolumn
\begin{tcblisting}{
    enhanced,                  % 开启增强绘图
    width=\columnwidth,
    title={Answer Example}, % 标题
    colframe=headerblue,       % 边框颜色
    colback=white,             % 背景颜色
    coltitle=white,            % 标题文字颜色
    fonttitle=\bfseries\Huge,
    arc=6mm,                   % 顶部圆角程度
    sharp corners=south,       % 【关键】底部设为直角
    boxrule=1.5pt,             % 边框粗细
    left=6mm, right=6mm, top=6mm, bottom=6mm, 
    toptitle=3mm, bottomtitle=3mm,
    listing only,              % 告诉它里面全是代码
    breakable,                 % 允许跨页
    listing options={          % 代码设置
        basicstyle=\small\ttfamily, 
        breaklines=true,       
        columns=fullflexible,  
        keepspaces=true,       
        language=Java,         
        tabsize=2,             
        showstringspaces=false 
    }
}
{
  "query_uid": "Txxxx"/"Gxx-x",
  "itinerary": {
    "summary": {
      "total_days": 天数,
      "total_travelers": 人数,
      "total_budget": 总预算,
      "calculated_total_cost": 计算总花费,
      "is_within_budget": true/false
    },
    "accommodation": {
      "hotel_name": "酒店名称",
      "room_type": [
        {
          "type": "房型（从单人房、大床房、双人房、家庭房中选择）",
          "quantity": "房间数量",
          "price_per_night": "每晚单价",
          "nights": "入住晚数"
        }
        // 若有多个房型，可继续添加
      ],
      "total_cost": 住宿总费用
    },
    "intercity_transport": {
      "transport_type": [
        {
          "description": "行程描述",
          "start_time": "HH:MM",
          "end_time": "HH:MM",
          "location_name": "目的地站点",
          "cost": "行程费用",
          "transportation_to": "高铁/飞机/动车/快速/特快/直达特快",
          "transportation_cost": "交通费用",
          "details": {
            "transport_number": "车次/航班编号",
            "price": "车票单价",
            "number": "车票数量"
          }
        },
        // 去程和返程
      ],
      "total_cost": 城际交通总费用
    },
    "daily_plans": [
      {
        "day": 1,
        "date": "YYYY-MM-DD",
        "starting_point": "起点",
        "ending_point": {
          "type": ""intercity_transport"/"accommodation"",
          "description": "今日终点描述",
          "start_time": "HH:MM",
          "end_time": "HH:MM",
          "location_name": "终点名称（车站或酒店）",
          "cost": 0,
          "transportation_to": "交通方式",
          "transportation_cost": 交通费用,
          "details": {
            "transport_time": "交通预计时间（HH:MM）",
            "transport_number": "车次/航班编号（仅城际交通时填写）",
            "line": "地铁线路（仅交通方式为地铁时填写）",
            "bus_number": "公交车号（仅交通方式为公交时填写）",
            "load_limit": 限载人数（仅打车或包车时填写）",
            "car_number": 车辆数（仅打车或包车时填写）
          }
        },
        "activities": [
          {
            "type": "intercity_transport"/"attraction"/"meal"/"accommodation"/"accommodation_check_in"/"accommodation_check_out",
            "description": "活动描述",
            "start_time": "HH:MM",
            "end_time": "HH:MM",
            "location_name": "地点名称",
            "cost": 活动费用（仅POI本身的费用，不包括交通费用）,
            "transportation_to": "交通方式",
            "transportation_cost": 交通费用,
            "details": {
              "transport_time": "交通预计时间（HH:MM）",
              // 根据类型填充详细信息
              "transport_number": "车次/航班编号（仅城际交通时填写）",
              "ticket_type": "票务类型（如观光票、一日票等，仅景点时填写）",
              "ticket_price": 票务单价（仅景点时填写）,
              "ticket_number": 票务数量（仅景点时填写）,
              "cuisine": "推荐菜品（仅餐饮时填写）",
              "line": "地铁线路（仅地铁时填写）",
              "bus_number": "公交车号（仅公交时填写）",
              "load_limit": 限载人数（仅打车或包车时填写）",
              "car_number": 车辆数（仅打车或包车时填写）
            }
          }
          // 每日的活动列表
        ]
      }
      // 每日计划列表
    ],
    "cost_breakdown": {
      "attractions": 景点总花费,
      "intercity_transportation": 城际交通总花费,
      "intracity_transportation": 市内交通总花费,
      "accommodation": 住宿总花费,
      "meals": 餐饮总花费,
      "other": 其他花费,
      "total": 总计
    }
  }
}
\end{tcblisting}

\newpage


\begin{promptbox}{User Query Generation Prompt}

\textbf{[Role: Data Generation Expert]} \\
You are a rigorous data generation expert specializing in creating structured user query data for Travel Planning AI benchmarks.

\sectiontitle{Task Description}
Generate a JSON dataset containing simulated user travel planning requests based on the strict rules below.

\sectiontitle{I. Core Rules}
\begin{enumerate}
    \item \textbf{Output Format:} You must output a complete JSON object following the structure below. Do not include any other text.
    
    \item \textbf{Data Structure:} Strictly adhere to the following JSON schema:
    \begin{tcolorbox}[enhanced, breakable, colback=white, colframe=black!30, boxrule=0.5pt]
    \begin{lstlisting}[style=plainjson]
{
  "version": "1.0",
  "description": "TravelEval Benchmark Dataset",
  "queries": [
    {
      "uid": "T0001",
      "tag": "easy",
      "start_city": "beijing",
      "target_city": "shanghai",
      "days": 3,
      "people_number": 4,
      "people_composition": {
        "adults": 2,
        "children": 1,
        "seniors": 1
      },
      "budget": 6800,
      "dates": "2024-10-01 to 2024-10-03",
      "transportation": {},
      "accommodations": {},
      "diet": {},
      "attractions": {},
      "rhythm": {},
      "nature_language": "我们一家四口（2大1小1老）想从北京去上海玩3天，预算6800元。",
      "nature_language_en": "Our family of four (2 adults, 1 child, 1 senior) wants to
      travel from Beijing to Shanghai for 3 days with a budget of 6800 RMB."
    }
  ]
}
    \end{lstlisting}
    \end{tcolorbox}
    \item \textbf{UID Generation:} \texttt{uid} generates sequentially starting from "T0001".
    
    \item \textbf{Field Constraints \& Sets:}
    \begin{itemize}
        \item \texttt{tag}: ["easy", "medium", "hard"]
        \item \texttt{start\_city}, \texttt{target\_city}: ["beijing", "chengdu", "chongqing", "guangzhou", "hangzhou", "nanjing", "shanghai", "shenzhen", "suzhou", "wuhan"]. (Note: Start and Target cities must not be the same).
        \item \texttt{days}: [2, 3, 4, 5, 6, 7]
        \item \texttt{people\_number}: [1, 2, 3, 4, 5, 6]
        \item \texttt{people\_composition}: Must satisfy \texttt{adults + children + seniors = people\_number}, with \texttt{adults} $\ge$ 1.
    \end{itemize}
\end{enumerate}

\sectiontitle{II. People Composition Rules}
\begin{itemize}
    \item \textbf{Solo:} \texttt{\{"adults": 1, "children": 0, "seniors": 0\}}
    \item \textbf{Duo:} Randomly select one:
    \begin{itemize}
        \item \texttt{\{"adults": 2, "children": 0, "seniors": 0\}} (Friends/Couple)
        \item \texttt{\{"adults": 1, "children": 1, "seniors": 0\}} (Parent \& Child)
        \item \texttt{\{"adults": 1, "children": 0, "seniors": 1\}} (Child \& Senior)
    \end{itemize}
    \item \textbf{Group (3-6):} Random generation satisfying: \texttt{adults} $\ge$ 1, \texttt{children} $\le$ 3, \texttt{seniors} $\le$ 2.
\end{itemize}

\sectiontitle{III. Budget Generation Rules}
Calculate \texttt{budget} strictly using these steps:
\begin{enumerate}
    \item \textbf{Step 1: Determine Consumption Class.} Randomly select a class with daily cost ($m$) and round-trip transport ($n$):
    \begin{itemize}
        \item \textit{Economy:} $m \in [100, 500]$, $n = 500$.
        \item \textit{Medium:} $m \in [500, 1000]$, $n = 1200$.
        \item \textit{High:} $m \in [1000, 5000]$, $n = 2000$.
    \end{itemize}
    \item \textbf{Step 2: Calculate Raw Budget.} \\
    \texttt{raw\_budget = (m * days + n) * people\_number}
    \item \textbf{Step 3: Rounding.}
    \begin{itemize}
        \item If \texttt{raw\_budget < 10000}: Round to the nearest multiple of 500.
        \item If \texttt{raw\_budget >= 10000}: Round to the nearest multiple of 2000.
    \end{itemize}
\end{enumerate}

\sectiontitle{IV. Difficulty level and preference generation rules}
User preference fields (\texttt{transportation}, \texttt{accommodations}, \texttt{diet}, \texttt{attractions}, \texttt{rhythm}) must be dictionary objects containing \texttt{"preferences"} (wants) and \texttt{"constraints"} (avoids) as string lists.

\vspace{0.2cm}
\textbf{Preference Tag Sets (Select from these exact Chinese terms):}

\begin{description}[font=\sffamily\bfseries\color{accent}]
    \item[transportation] ["飞机", "高铁", "自驾", "骑行"]
    
    \item[accommodations] ["管家服务", "桑拿", "家庭房", "SPA", "行政酒廊", "泳池", "智能客控", "免费停车", "山景房", "茶室", "停车场", "自营亲子房", "窗外好景", "日光浴场", "机器人服务", "Boss推荐", "网红泳池", "动人夜景", "私人泳池", "江河景房", "棋牌室", "私汤房", "充电桩", "酒店公寓", "影音房", "亲子主题房", "空气净化器", "多功能厅", "民宿", "智能马桶", "情侣房", "儿童俱乐部", "健身室", "24小时前台", "儿童乐园", "湖景房", "温泉", "拍照出片", "洗衣房", "设计师酒店", "会议厅", "四合院", "套房", "桌球室", "洗衣服务", "行李寄存", "提前入园", "美食酒店", "温泉泡汤", "电竞房", "空调", "商务中心", "穿梭机场班车", "洗衣机", "小而美", "别墅", "湖畔美居", "中式庭院", "历史名宅", "园林建筑", "钓鱼", "客栈", "特色住宿", "自营影音房", "电竞酒店", "老洋房", "厨房", "海景房", "迷人海景", "农家乐", "自营舒睡房", "位置超好", "儿童泳池", "宠物友好"]
    
    \item[diet]\hfill \\ \relax ["小吃", "面包甜点", "快餐简餐", "烧烤", "茶馆/茶室", "素食", "咖啡店", "粤菜", "湖北菜", "日本料理", "韩国料理", "清真菜", "西餐", "火锅", "本帮菜", "江浙菜", "川菜", "西北菜", "东南亚菜", "新疆菜", "其他中餐", "创意菜", "湘菜", "北京菜", "农家菜", "台湾菜", "海鲜", "客家菜", "酒吧/酒馆", "东北菜", "其他", "自助餐", "融合菜", "云南菜", "徽菜", "鲁菜", "西藏菜", "中东料理", "闽菜", "海南菜", "拉美料理", "亚洲菜"]
    
    \item[attractions] ["网红打卡点", "沉浸式体验", "夜经济热点", "亲子友好", "小众秘境", "城市地标", "美食目的地", "节庆限定", "非遗体验", "宠物特色", "历史古迹", "博物馆/纪念馆", "自然风光", "人文景观", "大学校园", "美术馆/艺术馆", "红色景点", "游乐园/体育娱乐", "图书馆", "园林", "其它", "文化旅游区", "公园", "商业街区"]
    
    \item[rhythm] ["慢游", "特种兵式"]
\end{description}

\sectiontitle{Difficulty Rules}
\begin{itemize}
    \item \textbf{Easy:}
    \begin{itemize}
        \item Constraint: Only basic fields.
        \item Preferences: All 5 fields must be empty dictionaries \texttt{\{\}}.
    \end{itemize}
    \item \textbf{Medium:}
    \begin{itemize}
        \item Constraint: Randomly select 1-2 preference categories.
        \item Logic: Add 1-2 tags to \texttt{"preferences"} only. \texttt{"constraints"} must remain empty \texttt{[]}.
    \end{itemize}
    \item \textbf{Hard:}
    \begin{itemize}
        \item Constraint: Total active preference categories must be 3 or 4.
        \item Logic: Must include at least 1 negative constraint (add value to \texttt{"constraints"}).
        \item Budget: $\sim$50\% of cases must use "Economy" class budget to create tight constraints.
        \item \textit{Context Awareness:} If seniors/children are present, preferences must logically align (e.g., Seniors $\to$ Slow rhythm).
    \end{itemize}
\end{itemize}

\sectiontitle{Other Fields}
\begin{itemize}
    \item \texttt{dates}: Generate a reasonable recent date range (YYYY-MM-DD format) matching \texttt{days}.
    \item \texttt{nature\_language}: A fluent, natural Chinese query summary based on all structured data.
    \item \texttt{nature\_language\_en}: English translation of the user query.
\end{itemize}

\sectiontitle{V. Output Requirements}
Generate a JSON object containing exactly \textbf{12 records}:
\begin{itemize}
    \item \textbf{4 Easy} records
    \item \textbf{4 Medium} records
    \item \textbf{4 Hard} records
\end{itemize}

\vspace{0.3cm}
\textbf{Output the valid JSON strictly following the format.}
\end{promptbox}

\newpage

\begin{promptbox}{Progressive Difficulty Query Generation Prompt}

\textbf{[Role: Data Generation Expert]} \\
You are a rigorous data generation expert specializing in creating structured user query data for Travel Planning AI benchmarks.

\sectiontitle{Task Description}
Based on all the rules below, generate a JSON dataset containing **50 "Progressive Difficulty Query Groups"**. Each group contains 3 user queries sharing the same travel core but with progressive difficulty levels (easy, medium, hard).

\sectiontitle{ I. Core Rules}
\begin{enumerate}
    \item \textbf{Output Format:} You must output a complete JSON object adhering to the structure below. Do not include any other text.
    \item \textbf{Data Structure:}

\begin{tcolorbox}[enhanced, breakable, colback=white, colframe=black!30, boxrule=0.5pt]
\begin{lstlisting}[style=plainjson]
{
  "version": "1.0",
  "description": "TravelEval Progressive Difficulty Dataset",
  "query_groups": [
    [
      {
        "uid": "G01-1",
        "tag": "easy",
        "start_city": "beijing",
        "target_city": "shanghai",
        "days": 3,
        "people_number": 2,
        "people_composition": {"adults": 2, "children": 0, "seniors": 0 },
        "budget": 5000,
        "dates": "2024-11-01 to 2024-11-03",
        "transportation": {},
        "accommodations": {},
        "diet": {},
        "attractions": {},
        "rhythm": {},
        "nature_language": "我们两个人从北京去上海玩3天，预算5000元。",
        "nature_language_en": "Two of us want to travel from Beijing to 
        Shanghai for 3 days with a budget of 5000 RMB" },  
        {
        "uid": "G01-2",
        "tag": "medium" },   {
        "uid": "G01-3",
        "tag": "hard"
      }
    ]
  ]
}
\end{lstlisting}
\end{tcolorbox}

    \item \textbf{UID Generation:} Format is \texttt{"GXX-Y"}. \texttt{XX} is the group number (starting from 01); \texttt{Y} is the sequence within the group (1 to 3), corresponding to tags "easy", "medium", "hard".
\end{enumerate}

\sectiontitle{II. Shared Core Parameters Generation Rules}
The 3 records within the same group must share the following **completely identical** core parameters:

\begin{enumerate}
    \item \textbf{City Pair:}
    \texttt{start\_city} and \texttt{target\_city} must be randomly selected from the following set:
    ["beijing", "chengdu", "chongqing", "guangzhou", "hangzhou", "nanjing", "shanghai", "shenzhen", "suzhou", "wuhan"].
    \textit{Note: Start and Target cities must not be the same.}

    \item \textbf{Itinerary \& People:}
    \begin{itemize}
        \item \texttt{days}: Randomly select from [2, 3, 4, 5, 6, 7].
        \item \texttt{people\_number}: Randomly select from [1, 2, 3, 4, 5, 6].
        \item \texttt{people\_composition}: Must satisfy \texttt{adults + children + seniors = people\_number}, with \texttt{adults} $\ge$ 1.
        \begin{itemize}
            \item \textbf{Solo:} \texttt{\{"adults": 1, "children": 0, "seniors": 0\}}
            \item \textbf{Duo:} Randomly select one: \texttt{\{"adults": 2, ...\}} (Friends/Couple), \texttt{\{"adults": 1, "children": 1, ...\}} (Parent \& Child), \texttt{\{"adults": 1, "seniors": 1, ...\}} (Child \& Senior).
            \item \textbf{Group (3-6):} Random generation satisfying: \texttt{adults} $\ge$ 1, \texttt{children} $\le$ 3, \texttt{seniors} $\le$ 2.
        \end{itemize}
    \end{itemize}

    \item \textbf{Budget Baseline \& Calculation:}
    \begin{enumerate}
        \item \textbf{Determine Unified Baseline:} First, randomly select a consumption class for the entire group and determine its corresponding $m$ (daily cost per person) and $n$ (round-trip transport):
        \begin{itemize}
            \item \textit{Economy:} $m \in [100, 500]$, $n = 500$.
            \item \textit{Medium:} $m \in [500, 1000]$, $n = 1200$.
            \item \textit{High:} $m \in [1000, 5000]$, $n = 2000$.
        \end{itemize}
        \item \textbf{Calculate Group Budget:} \\
        \texttt{raw\_budget = (m * days + n) * people\_number}
        \item \textbf{Budget Rounding:}
        \begin{itemize}
            \item If \texttt{raw\_budget < 10000}, budget must be a multiple of 500.
            \item If \texttt{raw\_budget >= 10000}, budget must be a multiple of 2000.
        \end{itemize}
        \item \textbf{Hard Difficulty Budget Constraint:} When generating the data (specifically for approximately 50\% of the groups), you must select the "Economy" tier as the consumption class to create difficulty.
    \end{enumerate}
\end{enumerate}

\sectiontitle{III. Difficulty Levels \& Progressive Preference Rules}
User preference fields (\texttt{transportation}, \texttt{accommodations}, \texttt{diet}, \texttt{attractions}, \texttt{rhythm}) must be dictionary objects containing \texttt{"preferences"} (wants) and \texttt{"constraints"} (avoids). All tags must be selected from the following sets:

\vspace{0.2cm}
\textbf{Tag Sets (Select values ONLY from these lists):}
\begin{description}
    \item[transportation] ["飞机", "高铁", "自驾", "骑行"]
    \item[accommodations] ["管家服务", "桑拿", "家庭房", "SPA", "行政酒廊", "泳池", "智能客控", "免费停车", "山景房", "茶室", "停车场", "自营亲子房", "窗外好景", "日光浴场", "机器人服务", "Boss推荐", "网红泳池", "动人夜景", "私人泳池", "江河景房", "棋牌室", "私汤房", "充电桩", "酒店公寓", "影音房", "亲子主题房", "空气净化器", "多功能厅", "民宿", "智能马桶", "情侣房", "儿童俱乐部", "健身室", "24小时前台", "儿童乐园", "湖景房", "温泉", "拍照出片", "洗衣房", "设计师酒店", "会议厅", "四合院", "套房", "桌球室", "洗衣服务", "行李寄存", "提前入园", "美食酒店", "温泉泡汤", "电竞房", "空调", "商务中心", "穿梭机场班车", "洗衣机", "小而美", "别墅", "湖畔美居", "中式庭院", "历史名宅", "园林建筑", "钓鱼", "客栈", "特色住宿", "自营影音房", "电竞酒店", "老洋房", "厨房", "海景房", "迷人海景", "农家乐", "自营舒睡房", "位置超好", "儿童泳池", "宠物友好"]
    \item[diet]  \hfill \\ \relax["小吃", "面包甜点", "快餐简餐", "烧烤", "茶馆/茶室", "素食", "咖啡店", "粤菜", "湖北菜", "日本料理", "韩国料理", "清真菜", "西餐", "火锅", "本帮菜", "江浙菜", "川菜", "西北菜", "东南亚菜", "新疆菜", "其他中餐", "创意菜", "湘菜", "北京菜", "农家菜", "台湾菜", "海鲜", "客家菜", "酒吧/酒馆", "东北菜", "其他", "自助餐", "融合菜", "云南菜", "徽菜", "鲁菜", "西藏菜", "中东料理", "闽菜", "海南菜", "拉美料理", "亚洲菜"]
    \item[attractions] ["网红打卡点", "沉浸式体验", "夜经济热点", "亲子友好", "小众秘境", "城市地标", "美食目的地", "节庆限定", "非遗体验", "宠物特色", "历史古迹", "博物馆/纪念馆", "自然风光", "人文景观", "大学校园", "美术馆/艺术馆", "红色景点", "游乐园/体育娱乐", "图书馆", "园林", "其它", "文化旅游区", "公园", "商业街区"]
    \item[rhythm] ["慢游", "特种兵式"]
\end{description}

\vspace{0.2cm}
\textbf{Progression Rules (Within the same group):}
\begin{itemize}
    \item \textbf{GXX-1 (Easy):}
    \begin{itemize}
        \item All five preference fields must be empty dictionaries: \texttt{\{\}}.
    \end{itemize}
    \item \textbf{GXX-2 (Medium):}
    \begin{itemize}
        \item Based on Easy, randomly select 1 to 2 preference categories.
        \item In each selected category, add only 1 or 2 tags to the \texttt{"preferences"} list.
        \item The \texttt{"constraints"} list must remain empty \texttt{[]}.
    \end{itemize}
    \item \textbf{GXX-3 (Hard):}
    \begin{itemize}
        \item Based on Medium, make the total number of defined preference categories reach 3 or 4 (i.e., add 1-2 more defined categories on top of the previous ones).
        \item Must include at least 1 \textbf{exclusionary constraint} (i.e., specify at least 1 tag in the \texttt{"constraints"} list of a field).
        \item \textbf{Preference Rationality:} When seniors or children are present, preferences should be logically associated (e.g., if Seniors are present $\to$ prefer "慢游" (Slow Travel); if Children are present $\to$ prefer "亲子友好" (Child-friendly) attractions or "家庭房" (Family Room) accommodation).
    \end{itemize}
\end{itemize}

\sectiontitle{ IV. Other Fields}
\begin{itemize}
    \item \texttt{dates}: Generate a reasonable recent date range, format "YYYY-MM-DD to YYYY-MM-DD", matching \texttt{days}.
    \item \texttt{nature\_language}: Generate a fluent, natural Chinese user query based on all structured information above.
    \item \texttt{nature\_language\_en}: Generate the corresponding English user query.
\end{itemize}

\sectiontitle{ V. Output Instructions}
Please generate **50 such Progressive Difficulty Query Groups** (Total 150 records). Output the complete JSON object directly. Do not include any extra explanation. Ensure the JSON format is completely correct and parsable.

\end{promptbox}

\newpage

\begin{promptbox}{Direct Prompting}
\noindent You are a professional travel planner possessing a complete database of Points of Interest (POIs) for the target city.

\vspace{0.5em}
\noindent \textbf{\# Important Instructions}
\begin{enumerate}[topsep=0pt, itemsep=0pt, parsep=0pt, leftmargin=*]
    \item Strictly follow the specified JSON format for the output.
    \item Please use standard POI names whenever possible (e.g., The Palace Museum, Shanghai Disney Resort).
\end{enumerate}

\vspace{0.5em}
\noindent For this planning task, intercity transportation must utilize the following specified stations: \texttt{<STATION CONSTRAINTS>}

\vspace{0.5em}
\noindent \textbf{User Request:} \texttt{<USER QUERY>}

\vspace{0.5em}
\noindent Please begin your planning and ensure the final output is valid JSON.
\end{promptbox}

\vspace{0.5cm} 

\begin{promptbox}{Zero-shot CoT Prompt}
\noindent You are a professional travel planner possessing a complete database of Points of Interest (POIs) for the target city.

\vspace{0.5em}
\noindent \textbf{\# Important Instructions}
\begin{enumerate}[topsep=0pt, itemsep=0pt, parsep=0pt, leftmargin=*]
    \item Please use standard POI names whenever possible (e.g., The Palace Museum, Shanghai Disney Resort).
\end{enumerate}

\vspace{0.5em}
\noindent \textbf{\# Reasoning Instructions} \\
Please think strictly according to the following steps and describe your reasoning process in natural language \textbf{before} providing the final answer:

\begin{enumerate}[topsep=0pt, itemsep=0pt, parsep=0pt, leftmargin=*]
    \item \textbf{Requirement Analysis:} Clarify core needs, constraints (budget, time, people), and preferences.
    \item \textbf{Data Selection:} Select options from the database (POIs, hotels, transport) and briefly explain reasons.
    \item \textbf{Itinerary Scheduling:} Plan daily activities ensuring route rationality and no conflicts.
    \item \textbf{Budget Calculation:} Estimate costs and ensure they are within the total budget.
\end{enumerate}

\vspace{0.5em}
\noindent After completing the above thinking, output the final plan strictly in the specified JSON format.

\vspace{0.5em}
\noindent For this planning task, intercity transportation must utilize the following specified stations: \texttt{\{station\_constraints\}}

\vspace{0.5em}
\noindent \textbf{User Request:} \texttt{\{user\_query\}}

\vspace{0.5em}
\noindent Please show the reasoning process first, then output the final plan.
\end{promptbox}

\newpage 

\begin{promptbox}{ReAct/Reflection Prompt}

\noindent \textbf{[Role 1: Travel Planner]} \\
You are a professional travel planner possessing a complete database of Points of Interest (POIs) for the target city.

\vspace{0.5em}
\noindent \textbf{\# Important Instructions}
\begin{enumerate}[topsep=0pt, itemsep=0pt, parsep=0pt, leftmargin=*]
    \item Use standard POI names (e.g., The Palace Museum, Shanghai Disney Resort).
    \item \textbf{Strictly follow the multi-turn interaction format below (up to \{self.max\_reflections\} turns):}
    \begin{itemize}[leftmargin=1em, topsep=0pt, itemsep=0pt]
        \item \textbf{Thought:} Analyze the current planning status and identify specific issues to solve next.
        \item \textbf{Action:} Plan the key data or decision needed for the next step.
        \item \textbf{Observation:} Provide the solution or data estimation.
    \end{itemize}
    \item When all core issues are resolved, output the final plan.
\end{enumerate}

\vspace{0.5em}
\noindent Intercity transportation station constraints: \texttt{\{station\_constraints\}} \\
\textbf{User Request:} \texttt{\{user\_query\}}

\noindent Please begin your ReAct loop, and finally output valid JSON strictly following the format.

\vspace{0.5em}
\hrule
\vspace{0.5em}

\noindent \textbf{[Role 2: Evaluation Expert]} \\
You are a travel planning evaluation expert required to critically reflect on the following plan. Please strictly output in the original JSON structure:

\vspace{0.5em}
\noindent \textbf{\# Input Data:} \texttt{\{response\}}

\vspace{0.5em}
\noindent \textbf{\# Reflection Framework}
\begin{enumerate}[topsep=0pt, itemsep=0pt, parsep=0pt, leftmargin=*]
    \item \textbf{Strengths of the current plan}
    \item \textbf{Identified issues or potential risks} (Budget, Time, Experience)
    \item \textbf{Optimization Priorities} (Mandatory Ranking)
\end{enumerate}

\vspace{0.5em}
\noindent \textbf{\# Output Requirements} \\
After generating the reflection results, output the JSON plan strictly following the format.

\end{promptbox}

\end{document}